%% file: bmvc.tex
\documentclass[table]{bmvc2k}
\usepackage{array} 
\usepackage{bm}
\usepackage{amsfonts}
\usepackage{algorithm}
\usepackage{algorithmic}
\usepackage{threeparttable}
\usepackage{graphicx}
\usepackage{color}
\usepackage{colortbl}
\usepackage{arydshln}
\usepackage{enumitem}
\usepackage{multirow}

\title{AtomGS: Atomizing Gaussian Splatting for High-Fidelity Radiance Field}

\addauthor{Rong Liu}{roliu@ict.usc.edu}{1}
\addauthor{Rui Xu}{rxu@ict.usc.edu}{1}
\addauthor{Yue Hu}{yuhu@ict.usc.edu}{1}
\addauthor{Meida Chen}{mechen@ict.usc.edu}{1}
\addauthor{Andrew Feng}{feng@ict.usc.edu}{1}

\addinstitution{
University of Southern California\\
Institute for Creative Technologies\\
Los Angeles, CA, USA
}

\runninghead{Liu, Xu, Hu, Chen, Feng}{\href{https://rongliu-leo.github.io/AtomGS/}{AtomGS}}

\begin{document}

\maketitle
\begin{figure}[!h]
    \centering
    \includegraphics[width=.9\linewidth]{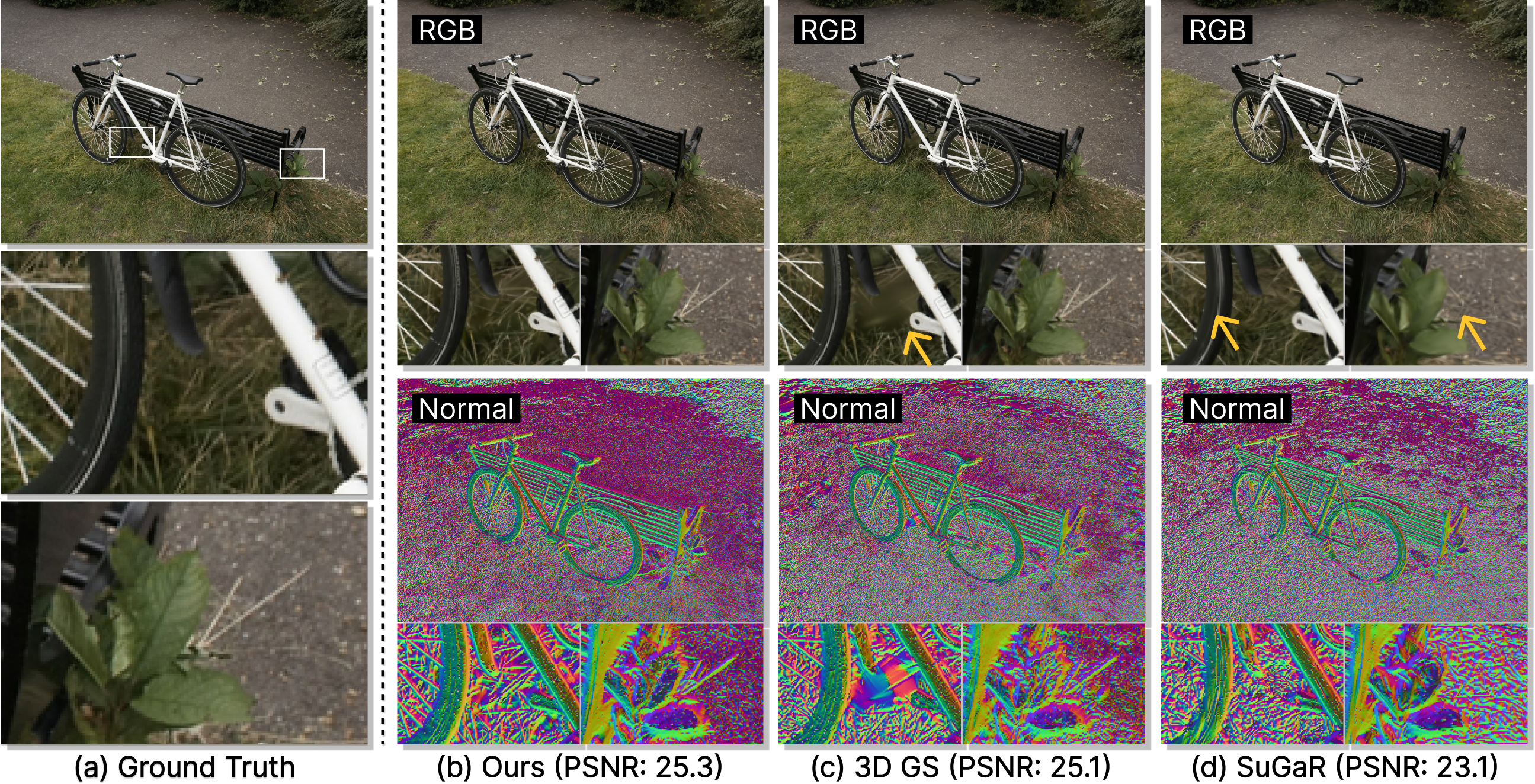}
    \caption{AtomGS outperforms existing methods in rendering quality and achieves competitive results in geometry accuracy by constraining Gaussians into Atom Gaussians and aligning them precisely with the natural geometry.}
    \label{fig:teaser}
\end{figure}
\vspace{-0.5cm}
\input{sec/0_abstract}
\input{sec/1_intro}

\input{sec/2_relatedwork}
\input{sec/3_method}

\input{sec/4_experiment}

\input{sec/5_conclusion}
\section*{Acknowledgments}
The authors would like to thank our primary sponsors of this research: Mr. Clayton Burford of the Battlespace Content Creation (BCC) team at Simulation and Training Technology Center (STTC). This work is supported by University Affiliated Research Center (UARC) award W911NF-14-D-0005. Statements and opinions expressed and content included do not necessarily reflect the position or the policy of the Government, and no official endorsement should be inferred.
\vspace{-0.2cm}
\bibliography{bmvc}

\newpage
\input{sup}

\end{document}

%% file: sec/0_abstract.tex
\begin{abstract}

3D Gaussian Splatting (3DGS) has recently advanced radiance field reconstruction by offering superior capabilities for novel view synthesis and real-time rendering speed. However, its strategy of blending optimization and adaptive density control might lead to sub-optimal results; it can sometimes yield noisy geometry and blurry artifacts due to prioritizing optimizing large Gaussians at the cost of adequately densifying smaller ones.
To address this, we introduce AtomGS, consisting of Atomized Proliferation and Geometry-Guided Optimization.
The Atomized Proliferation constrains ellipsoid Gaussians of various sizes into more uniform-sized Atom Gaussians. The strategy enhances the representation of areas with fine features by placing greater emphasis on densification in accordance with scene details. In addition, we proposed a Geometry-Guided Optimization approach that incorporates an Edge-Aware Normal Loss. This optimization method effectively smooths flat surfaces while preserving intricate details. 
Our evaluation shows that AtomGS outperforms existing state-of-the-art methods in rendering quality. Additionally, it achieves competitive accuracy in geometry reconstruction and offers a significant improvement in training speed over other SDF-based methods. 
More interactive demos can be found in our website (\href{https://rongliu-leo.github.io/AtomGS/}{https://rongliu-leo.github.io/AtomGS/}).

\end{abstract}

%% file: sec/1_intro.tex
\section{Introduction}
Multi-view 3D reconstruction and Novel view synthesis remain significant challenges in computer vision and graphics. A successful reconstruction requires both high-quality visual renderings from new viewpoints and precise capture of 3D geometry. These attributes ensure that models are visually appealing and accurate, making them valuable across various applications such as video games, VR/AR, digital twins, 3D mapping, simulation, scan-to-BIM, and more. Neural Radiance Fields (NeRF)~\cite{mildenhall2021nerf} has achieved significant progress in producing photorealistic renderings through implicit 3D representation; however, the limitation in rendering speed prevents its utilization in real-world applications. In response, 3D Gaussian Splatting (3DGS)~\cite{kerbl20233d} has emerged as a promising alternative, offering an explicit method that achieves fast rendering speeds while maintaining high rendering quality.

Existing works of 3DGS have primarily emphasized either improving rendering quality or enhancing 3D geometry accuracy~\cite{lu2023scaffoldgs,huang2024error,yan2023multi,guédon2023sugar,chen2023neusg}. Efforts that enhance rendering quality focus less on geometric precision, while those that aim to refine geometry often lead to reduced rendering quality. To address these challenges, we introduce AtomGS, an approach that enhances geometric precision in areas with fine details through our proposed Atomized Proliferation process and Edge-Aware Normal Loss. This enhancement in geometric detail consequently leads to improved rendering quality as shown in our experiments. Figure~\ref{fig:teaser} compares the results of our AtomGS method with existing methods, showcasing improvements in rendering quality and geometry surface normals.

\begin{figure}[!h]
\centering
  \begin{tabular}{cc}
    \includegraphics[width=0.45\linewidth]{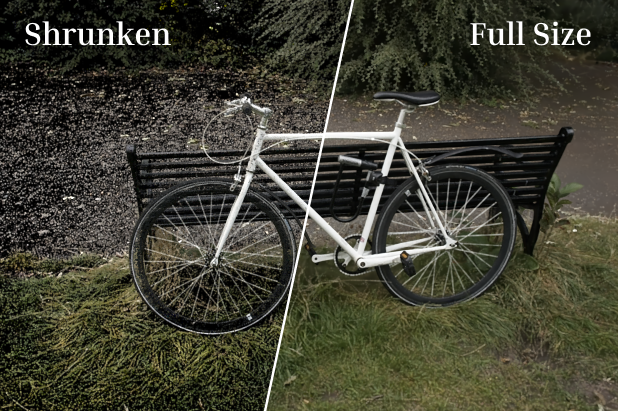} & 
    \includegraphics[width=0.45\linewidth]{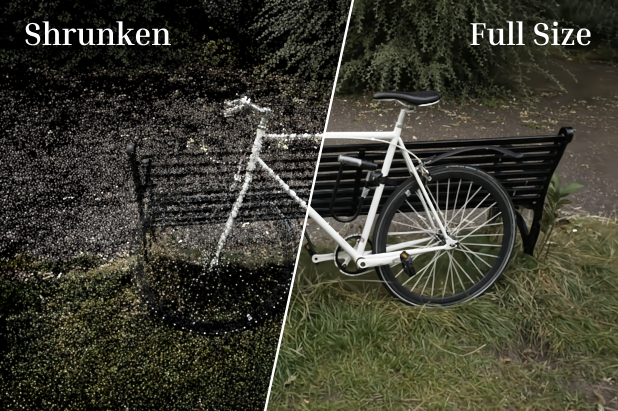}\\   
       (a) Ours  & (b) 3DGS
  \end{tabular}
    \caption{Rendering Comparison of 7k Results: Ours vs. 3DGS. We display images using both full-size and shrunken Gaussians, examining the rendering effects and Gaussian placements. Our approach results in more precise geometric alignments, visible in fine details like bicycle spokes and blades of grass.}
    \label{fig:proliferation}
\end{figure}

Our proposed AtomGS refines 3DGS by strategically deploying Atom Gaussians to ensure detailed coverage of complex scenes through the Atomized Proliferation process. In contrast to the 3DGS, AtomGS provides precise guidance on where to focus Gaussians for better 3D geometry optimization. Intuitively, smaller anisotropic Gaussians are constrained into uniformly-sized Atom Gaussians followed by a progressive split schedule to provide better coverage of scene details. Meanwhile, larger anisotropic Gaussians are retained to represent the background or geometric areas with few features. To facilitate this process, we have introduced an Edge-Aware Normal Loss that imposes stricter constraints on the positioning of Gaussians aligned with flat surfaces while allowing more flexibility for those on irregularly shaped areas. Specifically, we integrated weights derived from a 2D edge detector into the curvature map to compute this loss. In addition, during the subsequent pruning phase, Gaussians covering large, less detailed surfaces are merged, resulting in a similar or even reduced number of Gaussians compared to the original 3DGS while retaining competitive rendering quality. To summarize, the main contributions of our paper are listed as follows:

\begin{enumerate}[noitemsep,nolistsep]
\item We introduced an Atomized Proliferation strategy aimed at enhancing rendering quality by refining 3D geometric precision in areas with fine details.
\item We designed an Edge-Aware Normal Loss to enhance the reconstruction accuracy by preserving details in areas with irregular shapes while reducing noise on flat surfaces.
\item Our proposed AtomGS has achieved state-of-the-art performance on several benchmark datasets, excelling in both rendering quality and geometric precision.
\end{enumerate}






%% file: sec/2_relatedwork.tex
\section{Related work}
\subsection{Novel View Synthesis}

Volumetric methods, traditionally utilized for 3D scene representation~\cite{mildenhall2021nerf,wang2022clip,wang2021nerf,yen2021inerf,srinivasan2021nerv,deng2020nasa,jang2021codenerf,liu2021editing,noguchi2021neural}, involve subdividing space into discrete units known as voxels, allowing for detailed modeling of internal structures but often suffering from resolution limitations and high computational costs. 
In contrast, implicit neural representations~\cite{jiang2020local,wu2022object,ran2023neurar} have revolutionized 3D modeling by using continuous mathematical functions, learned by neural networks, to represent complex geometries and appearances without the need for spatial discretization. 

Building on these advancements, NeRF~\cite{mildenhall2021nerf} employ a coordinate-based neural network to encode volumetric scenes, providing unprecedented detail and realism in novel view synthesis, particularly effective in capturing complex light interactions and intricate details.
Building upon the original NeRF framework, InstantNGP~\cite{muller2022instant} introduces a multiresolution hash encoding that efficiently stores and retrieves neural network data, significantly boosting both training and inference speeds while striking a balance between performance and accuracy. 
On the other hand, Mip-NeRF 360~\cite{barron2022mipnerf} extends NeRF's capabilities to render large-scale, unbounded 360-degree scenes with consistent quality, effectively managing varied lighting conditions in expansive environments.




As we progress to alternative rendering solutions, 3DGS~\cite{kerbl20233d} has gained popularity for its ability to enhance visual rendering effects and speed through the optimized use of anisotropic 3D Gaussian ellipsoids and rasterized splatting techniques. 
Diverging from traditional 3DGS methods which freely drift and split, Lu et al.'s Scaffold-GS~\cite{lu2023scaffoldgs} leverages scene structure to guide the distribution of 3D Gaussians, allowing for adaptive modifications based on varying viewing angles and distances. 
Additionally, Huang et.al ~\cite{huang2024error} focuses on analyzing and reducing the artifacts caused by 3DGS errors, aiming to optimize rendering quality. 
For different levels of detail (LOD) in 3DGS scenes, Yan et.al, introduced a multi-scale approach~\cite{yan2023multi} to enable selective rendering that yields faster and more precise outcomes. 

While the methods above focus on enhancing the visual accuracy of rendered images, they often lack the geometric constraints necessary for high-quality surface reconstruction. In contrast, our method balances visual and geometric accuracy by refining the underlying geometry alignment of Gaussians.

\subsection{Multi-View 3D Mesh Reconstruction}
Inspired by NeRF~\cite{yen2021inerf}, NeuS~\cite{wang2021neus} integrates a Signed Distance Function (SDF) into radiance field to learn a neural SDF representation from multi-view images, thereby representing object surfaces with volumetric rendering accurately. Other concurrent works such as VolSDF~\cite{yariv2021volume} and UNISURF~\cite{oechsle2021unisurf} enhance surface reconstruction by improving ray sampling accuracy and simplifying the reconstruction process, respectively. Based on NeuS, Neuralangelo~\cite{li2023neuralangelo} proposes coarse-to-fine optimization on the hash grids and examines higher-order derivatives to reconstruct surfaces, leading to improved geometry accuracy and fine detail.

While SDF-based methods have greatly enhanced geometric surface reconstruction, they often result in poorer visual rendering performance and reduced reconstruction speeds due to the integration of SDF. Moreover, sphere initialization~\cite{Atzmon_2020_CVPR} is crucial for model convergence, limiting the application to datasets where the object is not centrally located.

Inspired by 3DGS, newer methods tackle precise geometric reconstruction with explicit representation. SuGaR~\cite{guédon2023sugar} proposes geometry constraints to regularize 3DGS, achieving geometry improvement, while NeuSG~\cite{chen2023neusg} introduces a scale regularizer to ensure the accuracy of the reconstructed surfaces by enforcing the 3D Gaussians to be extremely thin.

Despite significant progress in methods for accurate geometric reconstruction, they often lead to a reduction in rendering quality. Our method builds on better alignment between Gaussians and the inherent geometry and designs subsequent optimization processes for improved visual quality.

%% file: sec/3_method.tex
\section{Preliminary for 3D Gaussian Splatting}
Proposed as an alternative to NeRF-based methods, 3DGS combines differentiable optimization and non-differentiable adaptive density control for modeling the radiance field. 

\noindent\textbf{Ellipsoidal Gaussian Primitive:} A 3D Gaussian Primitive is defined as $G_i:=\{\bm{\mu},\alpha,\bm{q},\bm{s},\bm{f}\}$, where $\bm{\mu}\in \mathbb{R}^3$ represents the center of the Gaussian, also referred to as its position; $\alpha \in [0,1]$ signifies its opacity; $\bm{q}\in [-1,1]^4$ denotes the quaternion; $\bm{s}\in [0,\infty)^3$ stands for the scale; and $\bm{f} \in \mathbb{R}^d$ represents its learnable features. The color $\bm{c}$ is determined using spherical harmonics based on $\bm{f}$. The 3D covariance matrix is computed as $\bm{\Sigma}=\bm{R}\bm{S}\bm{S}^\top\bm{R}^\top$, where $\bm{R}$ and $\bm{S}$ are the rotation and scale matrices derived from $\bm{q}$ and $\bm{s}$, respectively.

During training, all Gaussian properties are optimized with the loss function: $\mathcal{L} = (1-\lambda)\mathcal{L}_1 + \lambda\mathcal{L}_{D-SSIM}$, where $\mathcal{L}_1 $ prioritizes pixel-wise accuracy, $\mathcal{L}_{D-SSIM}$ emphasizes structural similarity and perceptual quality, and $\lambda$ serves as the weighting factor.

\noindent\textbf{Adaptive Density Control}: The stage inserts, splits or prunes existing Gaussian primitives to better represent the 3D scene.
\begin{enumerate}[noitemsep,nolistsep]
    \item \textbf{SfM Initialization:} Given the SfM points, 3DGS calculates the mean distance to the closest three points, denoted as $\bm{d}\in [0,\infty)^i$, where $i$ is the number of initialized SfM points. Then it employs this distance to initialize isotropic Gaussians. Additionally, considering camera centers $\bm{\pi}\in \mathbb{R}^k$, where $k$ is the number of training camera poses, 3DGS determines its radius using $r = \max(\bm{\pi}-\Bar{\pi})$, where $\Bar{\pi}$ represents the mean of all camera centers. Subsequently, it sets the scale threshold to $\tau_s=0.01r$, which decides whether to clone or split the Gaussian if the gradient condition is satisfied.
    \item \textbf{Densification:} 3DGS adaptively densifies Gaussians to enhance scene detail capture. This densification process occurs regularly, targeting Gaussians with view-space positional gradients equal to or greater than the gradient threshold $\tau_p$. Following the gradient condition $\nabla_p\mathcal{L} \geq \tau_p$, it then checks if $||\bm{S}||_{\max}\geq\tau_s$. If the scale condition is met, the Gaussian is identified as over-reconstructed and split (creating two Gaussians with positions normally sampled based on the original). If not, it's classified as under-reconstructed, leading to the clone of an identical Gaussian.
    \item \textbf{Prune:} The point pruning phase involves removing redundant or less important Gaussians. This involves deleting Gaussians with the opacity $\alpha$ below a specified threshold. Moreover, to prevent producing noisy Gaussians near input cameras, the alpha values are gradually set closer to zero after a certain number of iterations. This adjustment facilitates the densification of necessary Gaussians while eliminating redundant ones.
\end{enumerate}

\noindent\textbf{Challenges:}
3DGS presents a hybrid optimization approach by integrating differentiable backpropagation with non-differentiable adaptive density control. However, it faces several challenges impacting its effectiveness. First, there's a lack of prioritization in the optimization process, where the method may focus on enlarging large Gaussians instead of densifying smaller ones to fill in gaps in geometry, or it might replicate transparent Gaussians to mimic a solid surface rather than enhancing the alignment and opacity of existing ones. Second, the absence of geometric regularization leads to misalignment of Gaussians with the underlying geometry, creating noisy artifacts that require opacity adjustments to clean up. Lastly, the simplistic approach to scale thresholding is influenced largely on camera pose radii than scene complexity, which restricts the method's ability to finely tune the splitting and cloning of Gaussians based on the detail needed for effective scene representation.
Consequently, while 3DGS produces a high-quality RGB radiance field, it may not adhere to the underlying geometric structures, which leads to noisy 3D mesh, blurry artifacts, and slower convergence speeds.

\section{Proposed Method}
To resolve the aforementioned issues, we propose a two-part approach. Firstly, Atomized Proliferation is introduced to enhance geometric precision in areas with intricate details, and secondly, a geometry-guided optimization is utilized to compactly modeling smooth surfaces while retaining enough primitives for fine details.

\subsection{Atomized Proliferation}
When handling SfM points, 3DGS alternates between densification and optimization to enhance scene representation. In contrast, our method initially constrains Gaussians that represent fine details into Atom Gaussians and prioritizes their proliferation to quickly align with the scene's inherent geometry. This is followed by a pruning strategy that merges the Gaussians representing large and smooth surfaces while preserving those representing detailed complexities. Figure~\ref{fig:proliferation} compares the resulted Gaussians at 7k iterations between ours and 3DGS.

\noindent\textbf{Atom Gaussian Primitive:}
Our process begins by analyzing the input SfM points to establish the Atom scale $\mathcal{S} = P_i(\bm{d})$, calculated from the $i$th percentile of ordered distances. This scale distinguishes between Gaussians capturing fine details (Atom Gaussians) and those covering broader background elements (traditional Gaussians). Atom Gaussians are distinct in being isotropic spheroids with a uniform size ($\bm{s}_1=\bm{s}_2=...=\bm{s}_n=(\mathcal{S})^3$), in contrast to traditional Gaussians, which are anisotropic ellipsoids of varying sizes. The uniform size of Atom Gaussians imposes the priority of densifying them to accurately fill gaps in the geometry and ensures a closer alignment with the actual 3D geometry of the scene, rather than optimizing elongated Gaussians that approximately cover these voids.

\noindent\textbf{Atomization:} This step checks the condition $||\bm{S}||_{\min}\leq\mathcal{S}$ regularly. If met, the Gaussian is designated as an Atom Gaussian with a size set to $\bm{s}=(\mathcal{S})^3$. Once categorized as Atom Gaussians, their scales are fixed and no longer optimized through backpropagation but through a geometric progression: $\mathcal{S}_{n} = \mathcal{S}_{n-1} \cdot r^{n-1}$, where $r = \sqrt[t_a]{p}$ and $t_a$ represents the total number of atomization iterations, and $p$ is the final proportion to the initial $\mathcal{S}$. This ensures that the atom scale decreases over iterations, progressively enhancing the representation of fine details.

\noindent\textbf{Densification:} 
This step is similar to original densification strategy, with the modification that allows larger Gaussians to be cloned. Moreover, we implement a linear warm-up approach to the split gradient threshold, enhancing the probability that a Gaussian will divide into smaller Atom Gaussians. Together with atomization, this strategy primarily aims to bridge the gaps in geometry.


\noindent\textbf{Prune:}
This step is similar to the low-opacity removal method but we increase the frequency of opacity resetting in the training. The focus is to merge Gaussians that depict large, simple surfaces, instead of eliminating the noisy ones near the camera. This step concludes the atom Gaussian strategy, allowing Gaussians to adapt their scales according to the complexity of the scenes. Through this refinement process, we retain Gaussians that capture intricate high-frequency details, while merging those that represent broad and smooth surfaces.

\subsection{Geometry-Guided Optimization}

\begin{figure}[!h]
    \centering
    \begin{tabular}{cc}
     \includegraphics[width=0.35\textwidth]{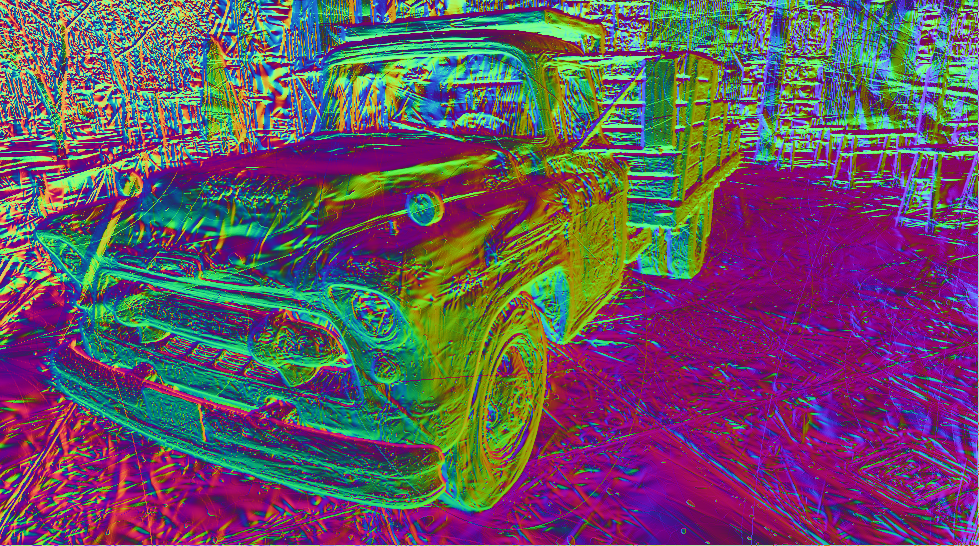}    
     &\includegraphics[width=0.35\textwidth]{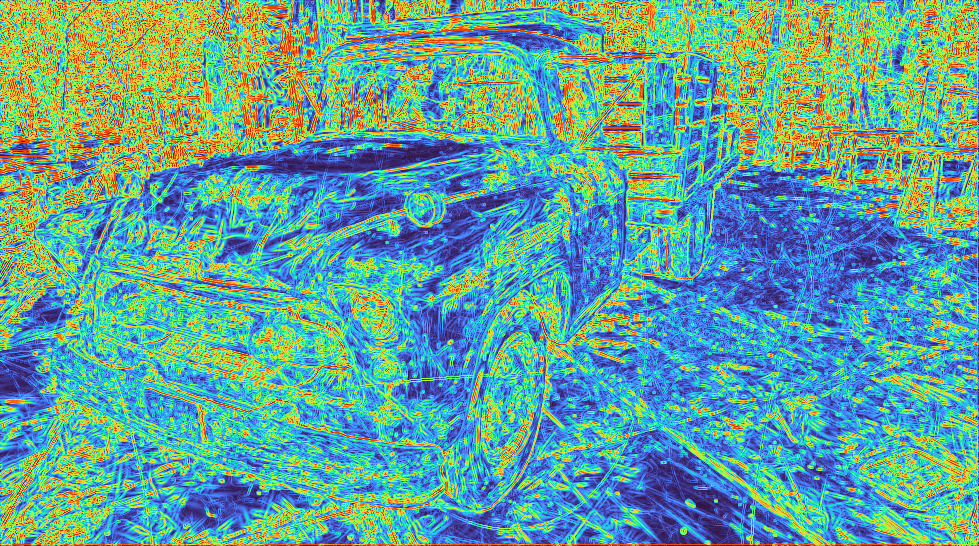}\\
     (a) Normal & (b) Curvature\\
     \includegraphics[width=0.35\textwidth]{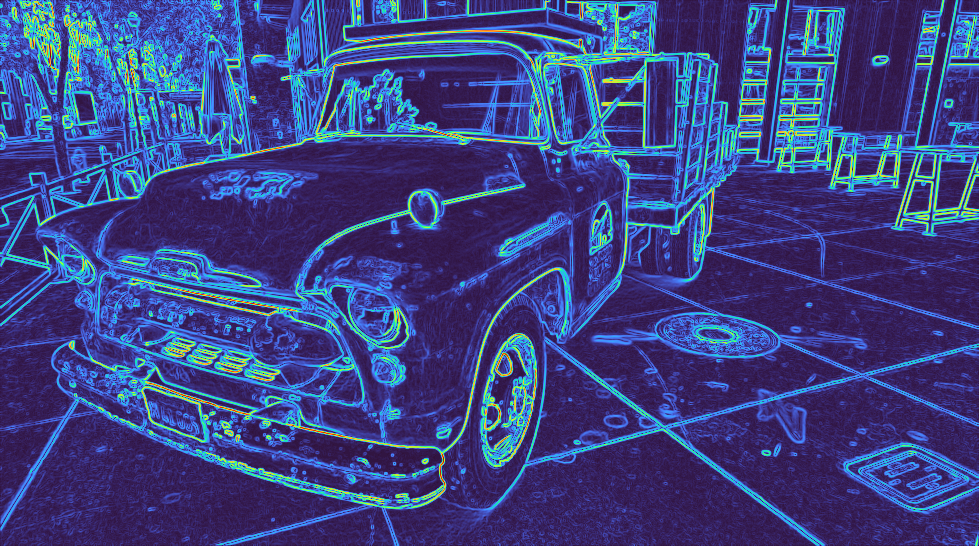} 
     &\includegraphics[width=0.35\textwidth]{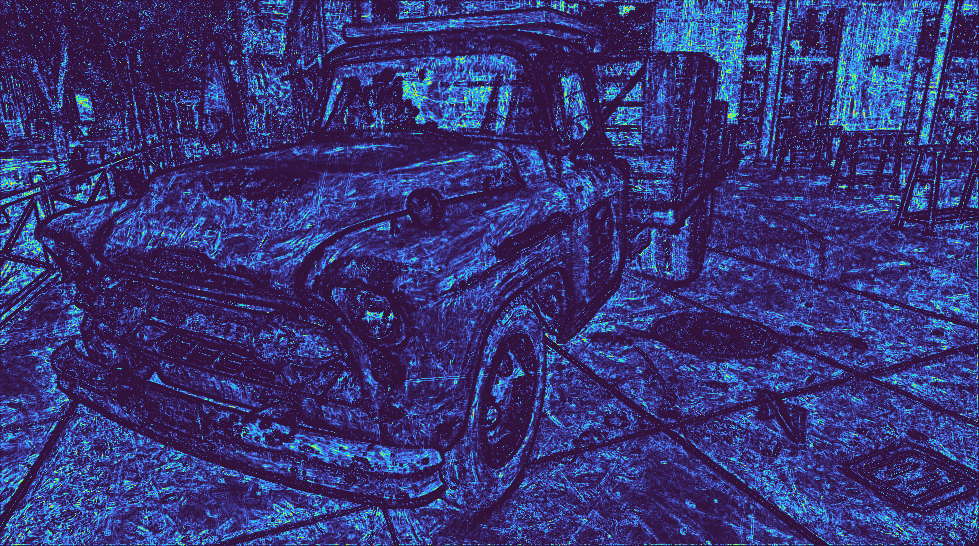} \\
       (c) Edge& (d) Edge-Aware Normal Loss
    \end{tabular}
    \caption{Illustration of the Edge-Aware Normal Loss. The image maps are rendered from the 3DGS 30k result without applying our proposed Edge-Aware Normal Loss. Loss map (d) highlights the areas where it can improve the geometry over the 3DGS result.}
    \label{fig:loss}
\end{figure}

To address the issue of Gaussians not always representing actual geometric structures we utilize a geometry-guided optimization, which comprises our proposed Edge-Aware Normal Loss and revised multi-scale SSIM loss. This optimization method ensures that enhancements focus on maintaining geometric accuracy without affecting the the RGB field fidelity. 

We first compute the normal map, $\bm{N}$, which visually represents the surface orientations derived from a radiance field's geometry (shown in Figure~\ref{fig:loss}a). This is calculated from an unprojected depth map, $\bm{D} \in \mathbb{R}^{H \times W \times 3}$, using the cross product of the depth map's gradients:
\begin{equation}
    \bm{N} = \frac{\nabla_x\bm{D}\times\nabla_y\bm{D}}{||\nabla_x\bm{D}\times\nabla_y\bm{D}||}
\end{equation}

Following this, the curvature map $|\nabla \bm{N}| \in [0,1]^{H \times W}$ is derived, representing the gradient magnitude of the normal map (shown in Figure~\ref{fig:loss}b). This map indicates the rate of change in surface normals, with higher values suggesting greater variability and lower values indicating smoothness. To enhance the geometric smoothness, one could optimize the curvature map. However, this might inadvertently smooth out high-frequency details such as sharp edges or fine structures, leading to an oversmoothed appearance.

To mitigate this, an edge map $|\nabla \bm{I}| \in [0,1]^{H \times W}$, derived from the gradient magnitude of the ground truth RGB image, is also computed (shown in Figure~\ref{fig:loss}c). The edge map helps preserve high-frequency details by excluding them from the smoothing process applied to the curvature map, thus maintaining essential geometric features on flat surfaces.

A weight function $\omega(x) = (x-1)^q$, where $q$ is a positive even integer, is introduced to finely balance the influence of the edge map on the curvature map. This weighting function is designed such that regions with low gradients (smooth areas) receive higher weights, promoting more smoothing, whereas regions with high gradients (sharp edges) receive lower weights to preserve detail. The tolerance $q$ determines the level of sensitivity to gradients, effectively controlling the extent to which details are either preserved or smoothed out.

The Edge-Aware Normal Loss (shown in Figure~\ref{fig:loss}d) is formulated as follows:

\begin{equation}
    \mathcal{L}_{normal} = \frac{1}{HW}\sum_i^H\sum_j^W|\nabla \bm{N}| \otimes \omega(|\nabla \bm{I}|) 
\end{equation}

For improved perceptual performance, we have replaced the SSIM loss with a Multi-Scale SSIM (MS-SSIM) loss \cite{wang2003multiscale} to capture a richer variety of camera view variations. The composite loss function is formulated as follows:
\begin{equation}
    \mathcal{L} = (1-\lambda_{ms-ssim})\mathcal{L}_1 + \lambda_{ms-ssim}\mathcal{L}_{ms-ssim}+\lambda_{normal}\mathcal{L}_{normal},
\end{equation}
where $\lambda_{ms-ssim}$ and $\lambda_{normal}$ are hyperparameters that determine the respective contributions of $\mathcal{L}_{ms-ssim}$ and $\mathcal{L}_{normal}$ to the overall loss function.


%% file: sec/4_experiment.tex
\section{Experiments}

\begin{figure}[!ht]
    \centering
    \includegraphics[width=0.9\textwidth]{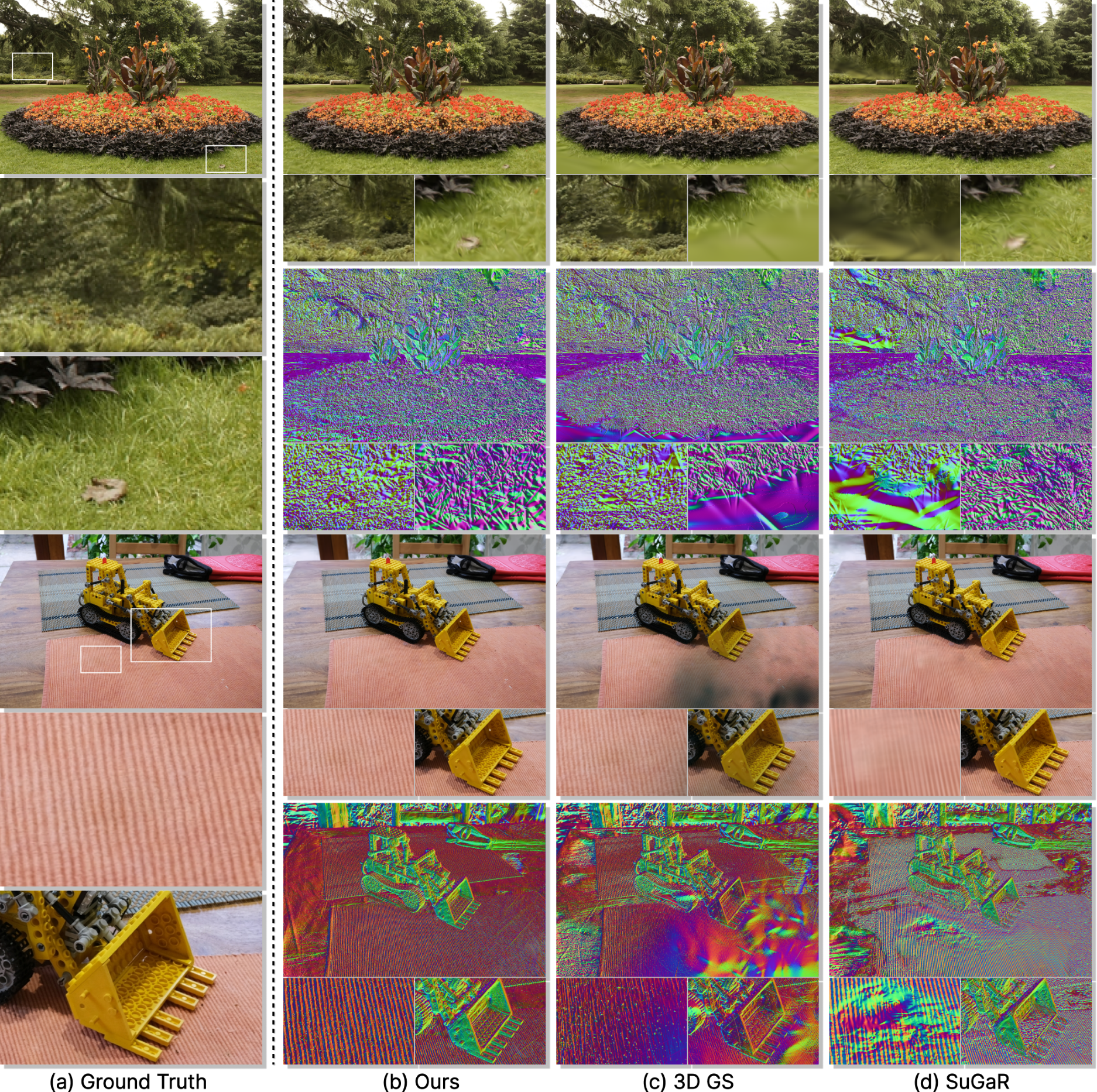}
    \caption{Radiance Field Comparison on the Mip-NeRF360 Dataset.}
    \label{fig:qualitative}
\end{figure}

\begin{table}[htbp]
    \centering
     \resizebox{0.6\linewidth}{!}{\begin{tabular}{cc|ccc|cccc}
    \hline
        
        & \multirow{2}{*}{Methods}& \multicolumn{3}{c|}{Mip-NeRF360} & \multicolumn{3}{c}{Tanks\&Temples}\\
         & &$\text{PSNR}^\uparrow$ &$\text{SSIM}^\uparrow$& $\text{LPIPS}^\downarrow$& $\text{PSNR}^\uparrow$& $\text{SSIM}^\uparrow$& $\text{LPIPS}^\downarrow$\\
         \hline
         
         \parbox[t]{2mm}{\multirow{4}{*}{\rotatebox[origin=c]{90}{non-explicit}}}
         
         & Plenoxels~\cite{fridovich2022plenoxels} & 23.08 &0.626 &0.463&21.08 &0.719&  0.379\\
         & Instant-NGP~\cite{muller2022instant} & 25.59 &0.699& 0.331&21.92&0.745 & 0.305\\ 
          &Mip-NeRF360~\cite{barron2022mipnerf}~ &\cellcolor{red!50}27.69&0.792& 0.237&22.22&0.759 & 0.257\\ 
        &TRIPS~\cite{franke2024trips}& 25.94& 0.772&0.233& \cellcolor{red!50}24.64& 0.808&0.213\\ \hdashline
         %

         \parbox[t]{2mm}{\multirow{5}{*}{\rotatebox[origin=c]{90}{explicit}}}
         
         & 3DGS~\cite{kerbl20233d} & \cellcolor{yellow!50}27.21 & \cellcolor{orange!50}0.815&\cellcolor{orange!50}0.214&23.14&\cellcolor{orange!50}0.841&\cellcolor{orange!50} 0.183\\
        &SuGaR~\cite{guédon2023sugar}& 25.51& 0.756 & 0.268& 22.68& 0.794&0.217\\
        &2DGS~\cite{huang20242d}&27.02&\cellcolor{yellow!50}0.804&\cellcolor{yellow!50}0.238&-&-&-\\

        &GES~\cite{hamdi2024ges}& 26.91 & 0.794 & 0.250&\cellcolor{yellow!50}23.35&\cellcolor{yellow!50}0.836&\cellcolor{yellow!50}0.198\\ 
        &AtomGS (Ours)&\cellcolor{orange!50} 27.38 &\cellcolor{red!50} 0.816&\cellcolor{red!50}0.211&\cellcolor{orange!50}23.70&\cellcolor{red!50}0.849&\cellcolor{red!50} 0.166 \\ \hline
    \end{tabular}}
    \caption{Quantitative evaluation of 2D rendering results on the Mip-NeRF360 and Tanks\&Temples datasets.}
    \vspace{-0.5cm}
    \label{tab:360dataset}
\end{table}

This section presents comprehensive evaluations of our designed AtomGS to compare its performance in both rendering quality and 3D geometry precision against previous state-of-the-art methods. 
In our quantitative tables, a dashed line distinguishes between methods that explicitly represent the scene as Gaussian primitives and non-explicit methods that encode appearance or geometric information within a network. We adapt the metrics from the original papers whenever possible for consistency and comparability.

\noindent\textbf{Datasets and Metrics:}
In our experiments, we evaluated the proposed AtomGS method using three datasets: Mip-NeRF360~\cite{barron2022mipnerf}, Tanks\&Temples~\cite{Knapitsch2017}, and DTU~\cite{aanaes2016large}. We assessed rendering quality using three commonly employed metrics—Peak Signal-to-Noise Ratio (PSNR), Structural Similarity Index Measure (SSIM)~\cite{1284395}, and Learned Perceptual Image Patch Similarity (LPIPS)~\cite{zhang2018perceptual}—on Mip-NeRF360 and Tanks\&Temples datasets. Additionally, we measured the geometry precision using the chamfer distance on the DTU dataset.

\noindent\textbf{2D Rendering Quality:}
Table~\ref{tab:360dataset} provides a quantitative evaluation of rendering quality for the selected methods. Our approach surpasses all other explicit methods in three key metrics and consistently maintains a top-two performance compared to all methods evaluated.
Figure~\ref{fig:qualitative} offers a qualitative comparison of our approach against two other explicit methods: 3DGS and SuGaR. In the flower scene, both 3DGS and SuGaR show varying levels of blurriness in areas with high frequency details, while our method maintains sharpness. In the kitchen scene, 3DGS presents noticeable artifacts close to the camera. Although SuGaR enhances surface smoothness and reduces noises, it causes unsightly distortions in the geometry of the table mat. In contrast, our method not only reproduces the smooth surface of the kitchen table but also preserves the intricate details of both the table mat and the Lego.

\begin{figure}[htbp]
    \centering
    \begin{tabular}{cccc}
         \rotatebox{90}{DTU 24}
         & \raisebox{-0.3\height}{\includegraphics[width=0.23\textwidth]{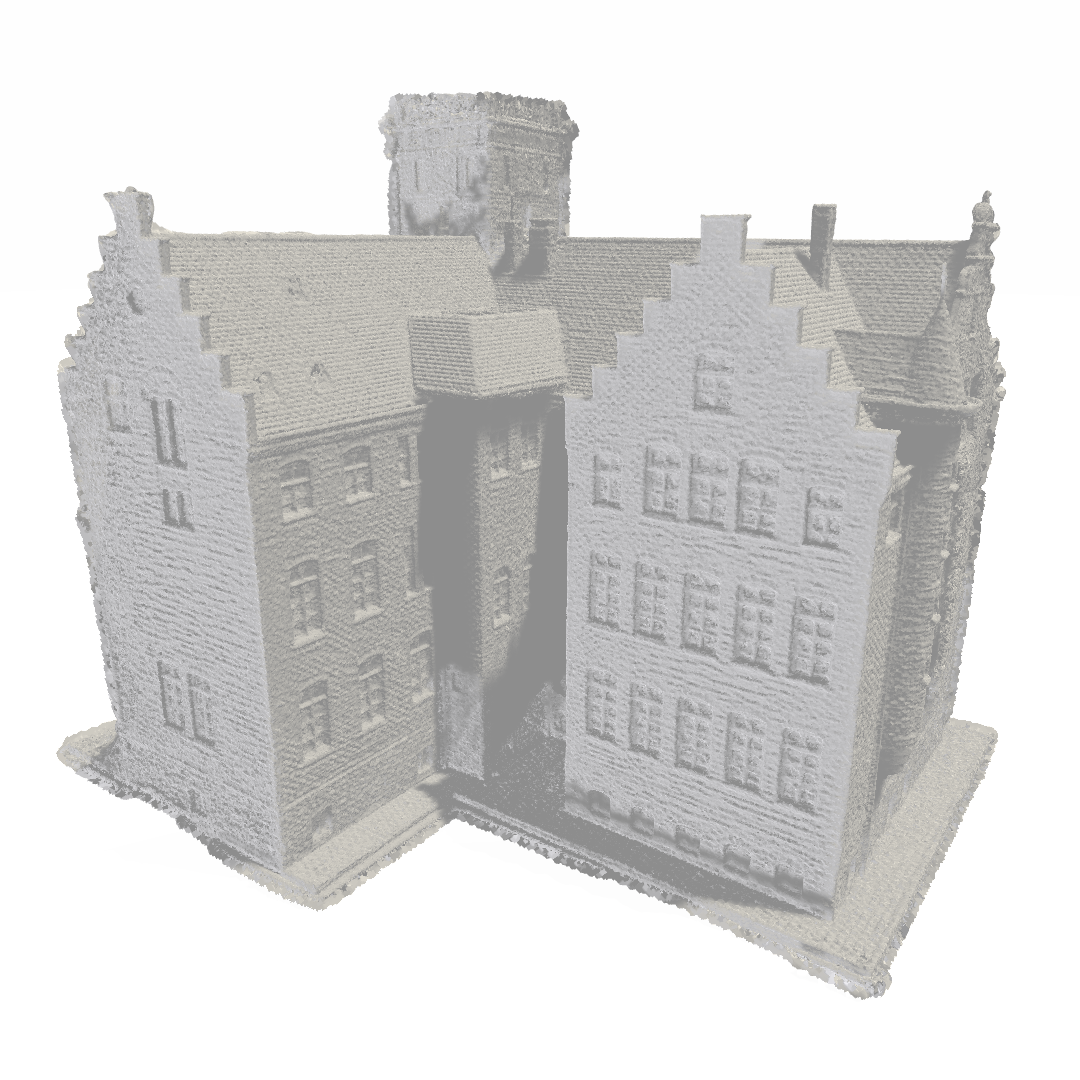}}
         & \raisebox{-0.3\height}{\includegraphics[width=0.23\textwidth]{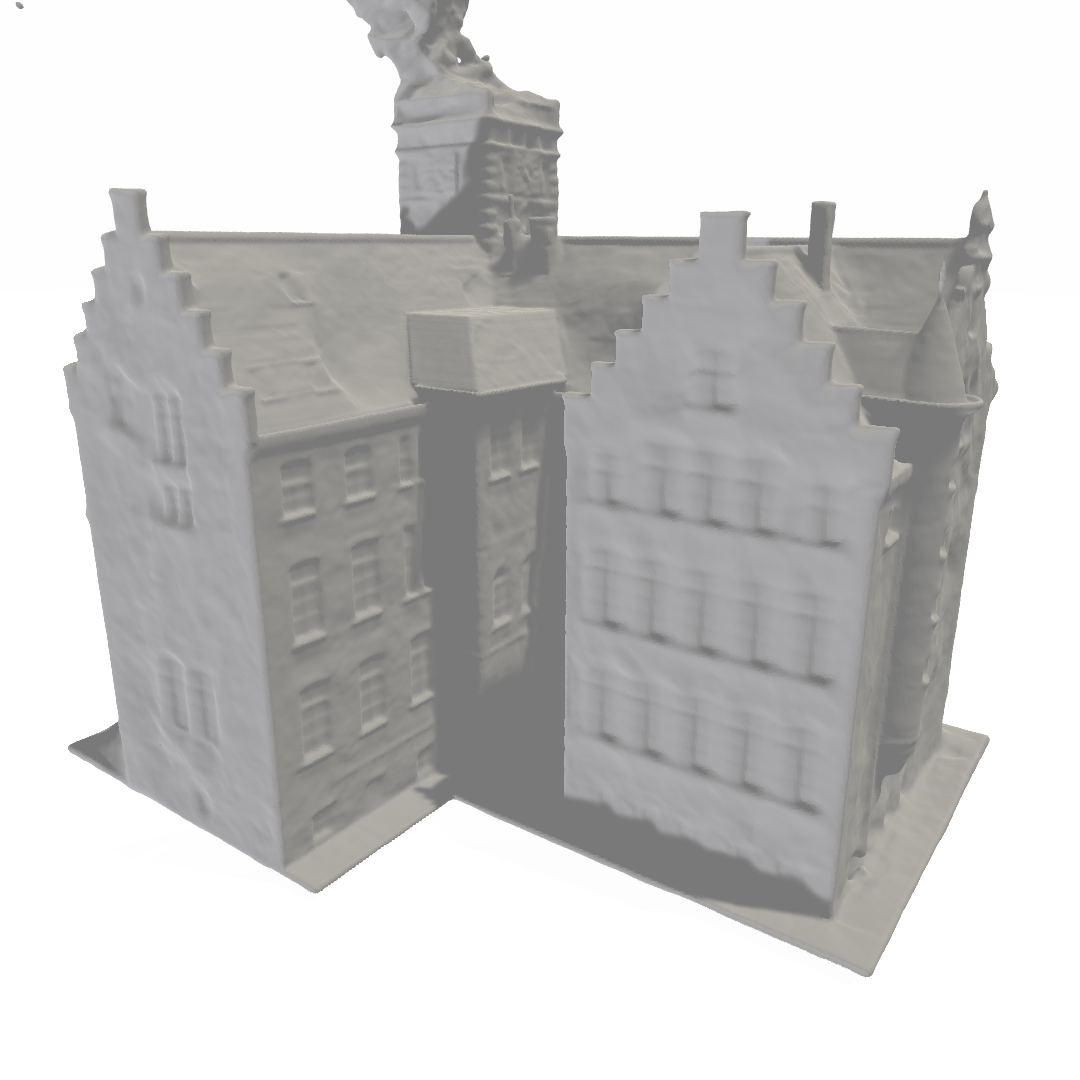}}
         & \raisebox{-0.3\height}{\includegraphics[width=0.23\textwidth]{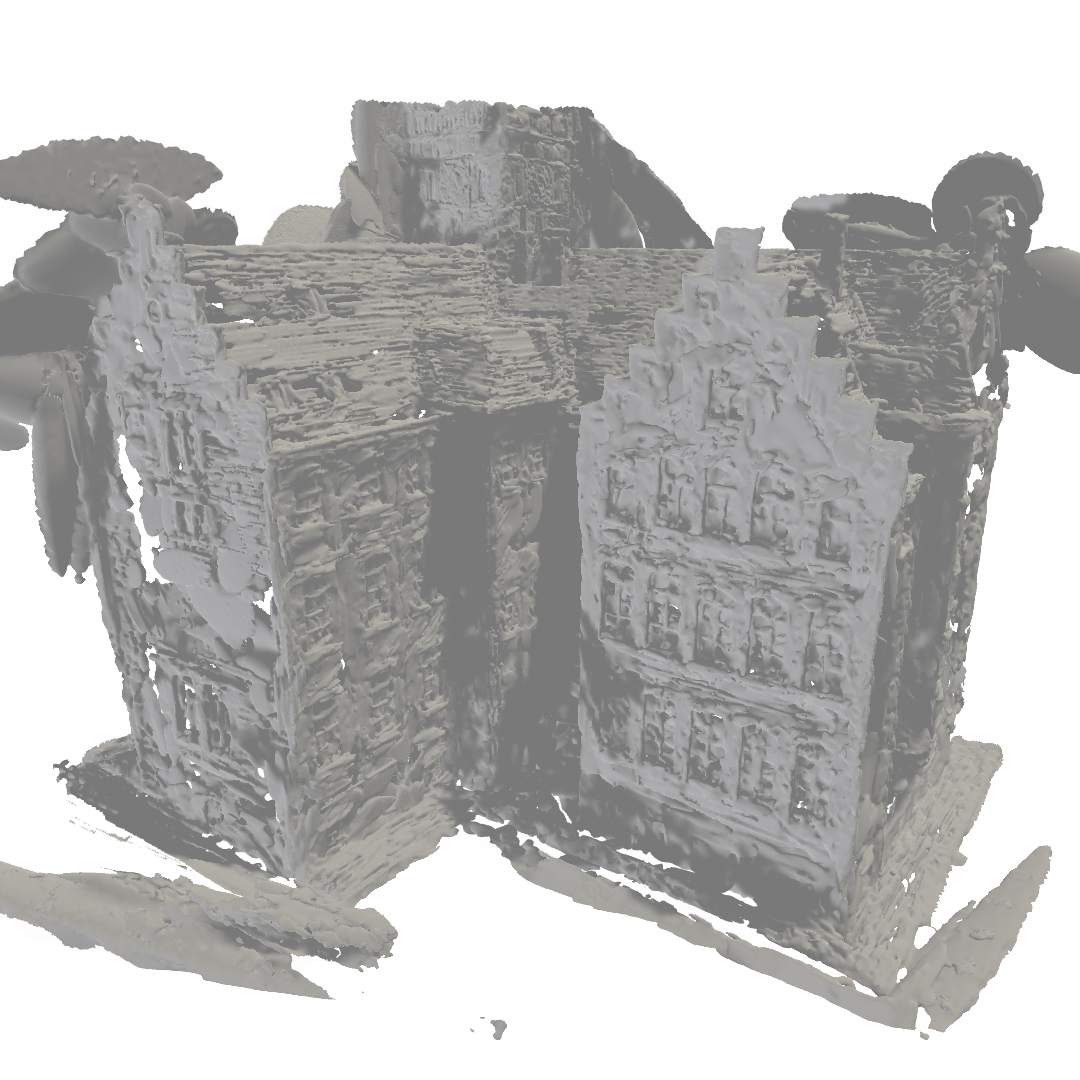}} \\

         \rotatebox{90}{Hotdog}
         & \raisebox{-0.3\height}{\includegraphics[width=0.23\textwidth]{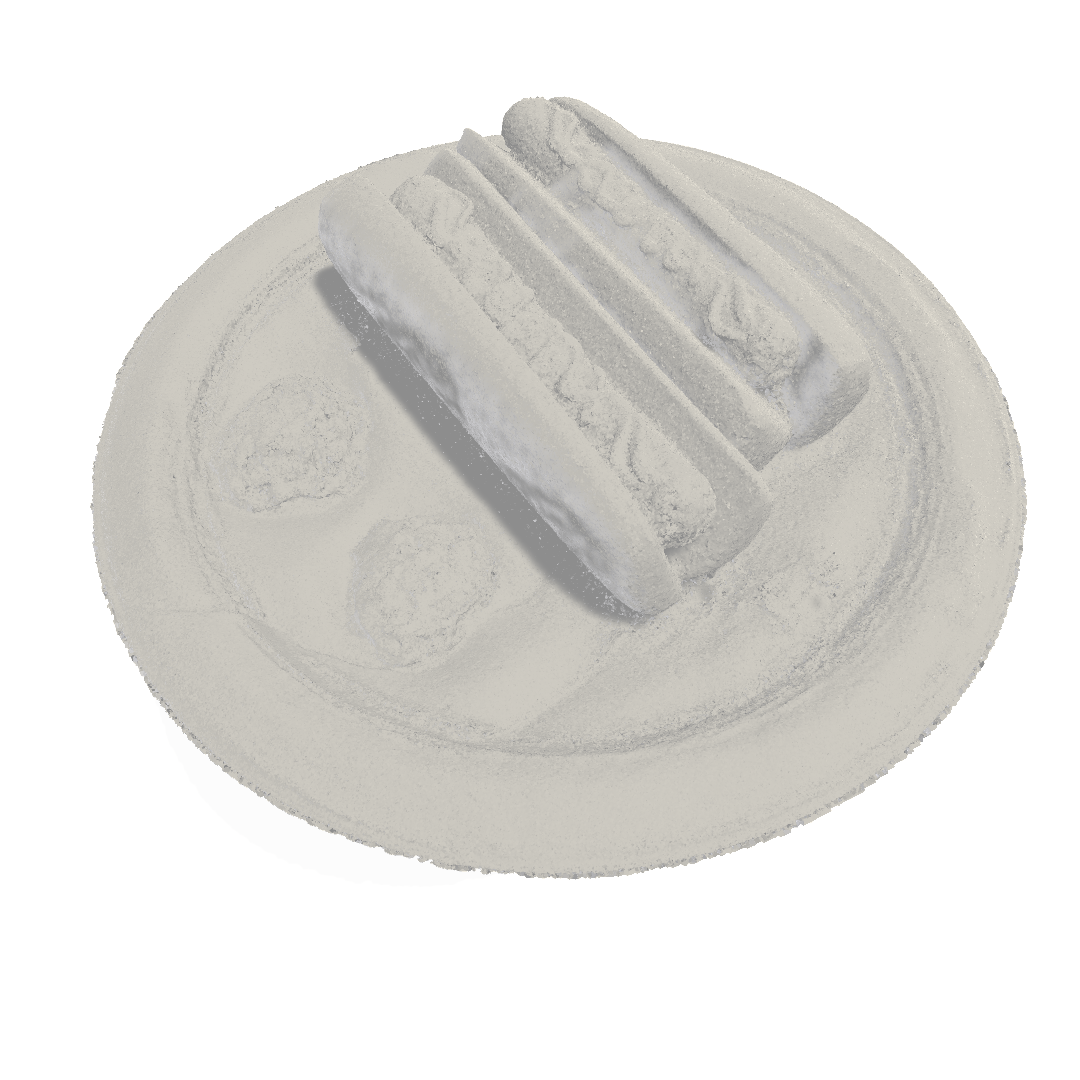}}
         & \raisebox{-0.3\height}{\includegraphics[width=0.23\textwidth]{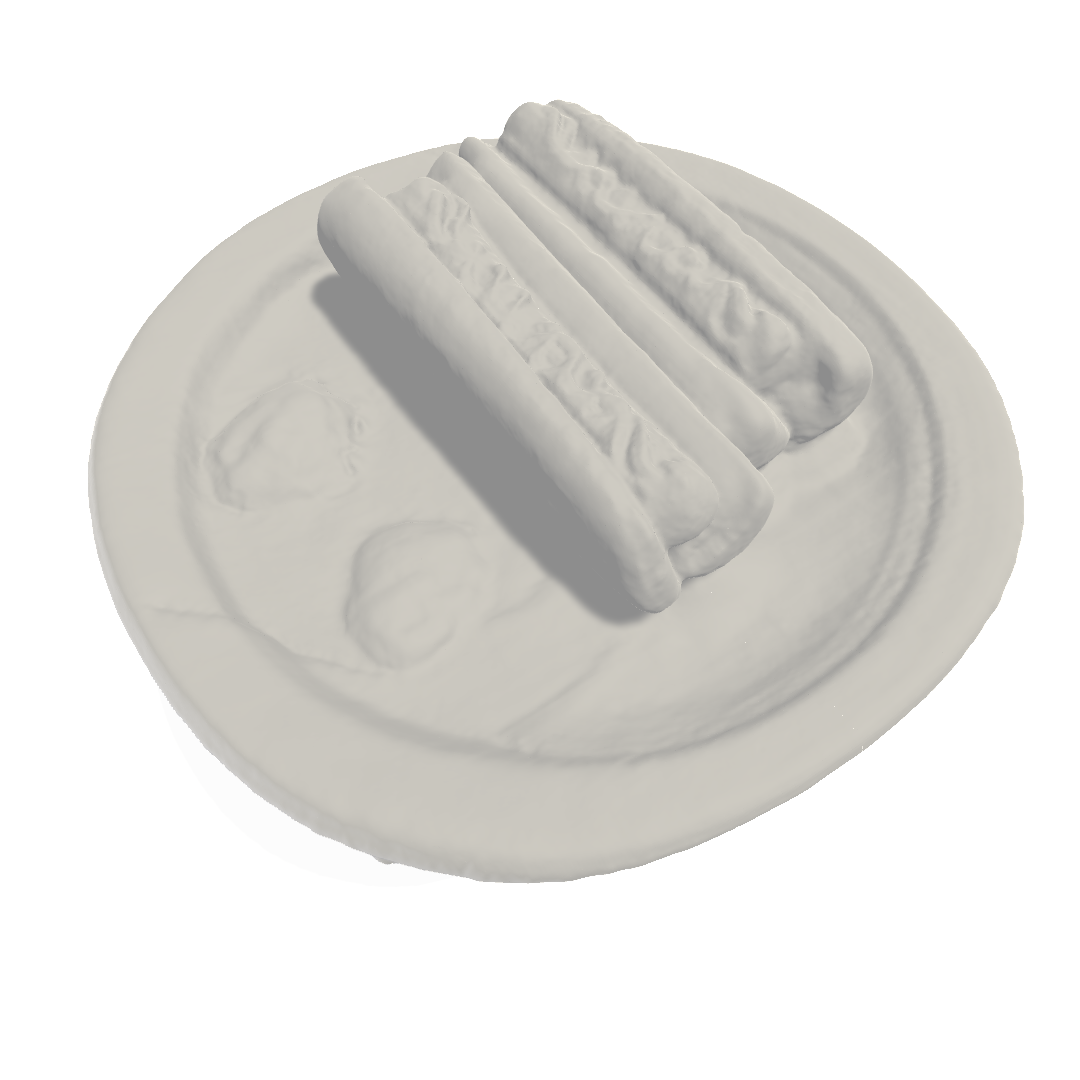}}
         & \raisebox{-0.3\height}{\includegraphics[width=0.23\textwidth]{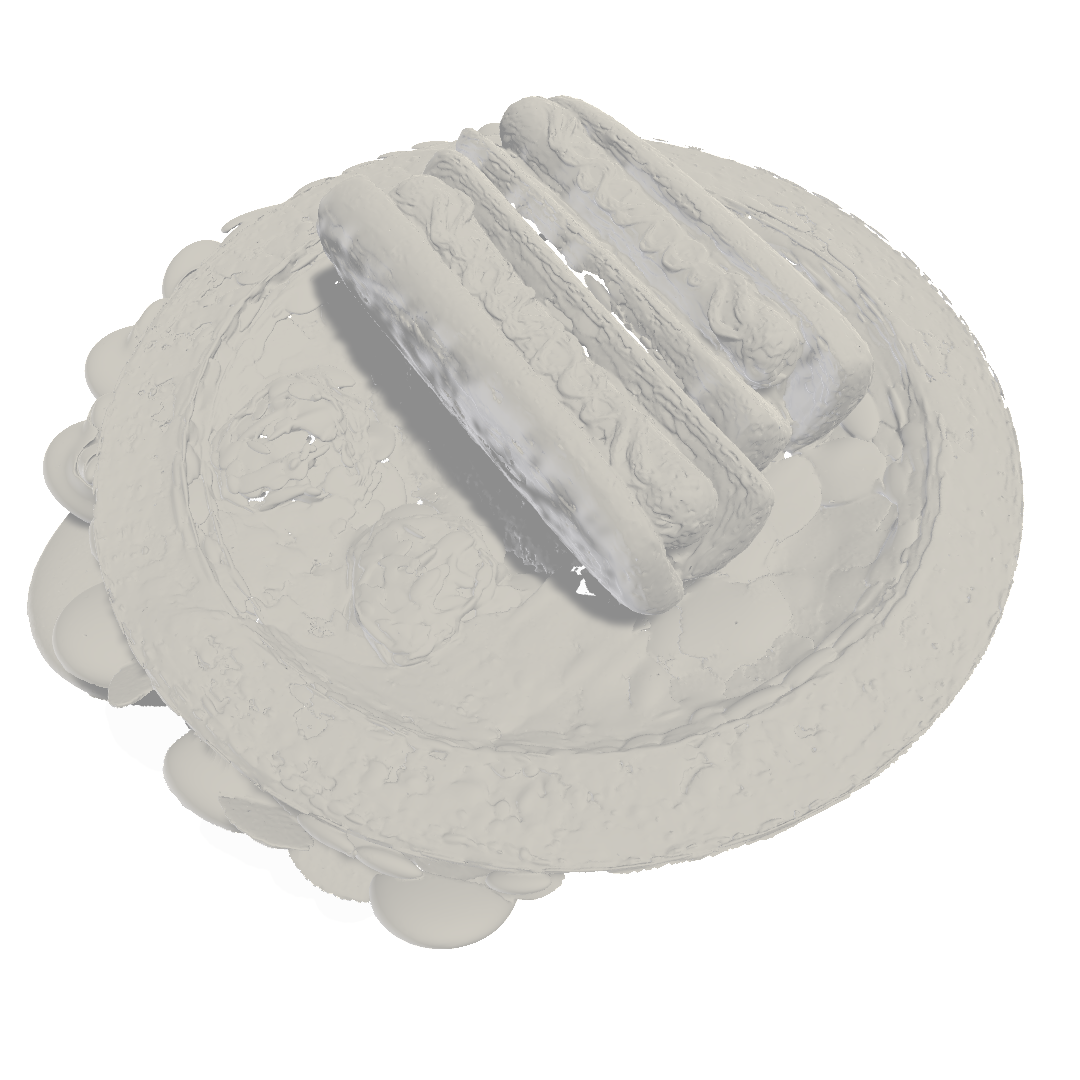}} \\
         & (a) Ours & (b) NeuS & (c) SuGaR \\
    \end{tabular}
    \caption{Mesh Comparison on the DTU and NeRF Synthetic Datasets~\cite{aanaes2016large, mildenhall2021nerf}}
    \vspace{-0.5cm}
    \label{fig:mesh}
\end{figure}

\begin{table}[htbp]
\centering
\resizebox{0.8\linewidth}{!}{\begin{tabular}{cc|ccccccccccccccc|cc}
\hline
 & Methods & 24 & 37 & 40 & 55 & 63 & 65 & 69 & 83 & 97 & 105 & 106 & 110 & 114 & 118 & 122 & $\text{Mean}^\downarrow$&Train$^\downarrow$ \\
\hline
\parbox[t]{2mm}{\multirow{4}{*}{\rotatebox[origin=c]{90}{implicit}}}
& NeRF~\cite{mildenhall2021nerf} & 1.90 & 1.60 & 1.85 & 0.58 & 2.28 & 1.27 & 1.47 & 1.67 & 2.05 & 1.07 & 0.88 & 2.53 & 1.06 & 1.15 & 0.96 & 1.49 & $\sim$4h \\
&VolSDF~\cite{yariv2021volume} & 1.14 & 1.26 & 0.81 & 0.49 & 1.25 & \cellcolor{yellow!50}0.7 & \cellcolor{yellow!50}0.72 & \cellcolor{orange!50}1.29 & 1.18 & \cellcolor{orange!50}0.7 & \cellcolor{yellow!50}0.66 & \cellcolor{orange!50}1.08 & 0.42 & 0.61 & 0.55 & 0.86 & $\sim$6h \\
&NeuS~\cite{wang2021neus} & 1.00 & 1.37 & 0.93 & 0.43 & 1.10 & \cellcolor{orange!50}0.65 & \cellcolor{orange!50}0.57 & 1.48 & \cellcolor{orange!50}1.09 & 0.83 & \cellcolor{orange!50}0.52 & \cellcolor{yellow!50}1.2 & \cellcolor{orange!50}0.35 & \cellcolor{orange!50}0.49 & 0.54 & 0.84 & $\sim$6h \\
&Neuralangelo~\cite{li2023neuralangelo} & \cellcolor{red!50}0.37 & \cellcolor{red!50}0.72 & \cellcolor{red!50}0.35 & \cellcolor{red!50}0.35 & \cellcolor{red!50}0.87 & \cellcolor{red!50}0.54 & \cellcolor{red!50}0.53 & \cellcolor{yellow!50}1.29 & \cellcolor{red!50}0.97 & \cellcolor{yellow!50}0.73 & \cellcolor{red!50}0.47 & \cellcolor{red!50}0.74 & \cellcolor{red!50}0.32 & \cellcolor{red!50}0.41 & \cellcolor{red!50}0.43 & \cellcolor{red!50}0.61 & $\sim$12h \\ \hdashline
\parbox[t]{2mm}{\multirow{4}{*}{\rotatebox[origin=c]{90}{explicit}}}
&3DGS~\cite{kerbl20233d} & 2.14 & 1.53 & 2.08 & 1.68 & 3.49 & 2.21 & 1.43 & 2.07 & 2.22 & 1.75 & 1.79 & 2.55 & 1.53 & 1.52 & 1.50 & 1.96 & \cellcolor{orange!50}0.19h \\
&SuGaR~\cite{guédon2023sugar} & 1.47 & 1.33 & 1.13 & 0.61 & 2.25 & 1.71 & 1.15 & 1.63 & 1.62 & 1.07 & 0.79 & 2.45 & 0.98 & 0.88 & 0.79 & 1.33 & 1.28h \\
&2DGS~\cite{huang20242d} & \cellcolor{orange!50}0.48 & \cellcolor{yellow!50}0.91 & \cellcolor{orange!50}0.39 & \cellcolor{orange!50}0.39 & \cellcolor{orange!50}1.01 & 0.83 & 0.81 & 1.36 & 1.27 & 0.76 & 0.7 & 1.40 & 0.40 & 0.76 & \cellcolor{yellow!50}0.52 & \cellcolor{yellow!50}0.80 & \cellcolor{yellow!50}0.31h \\
&AtomGS (Ours) & \cellcolor{yellow!50}0.51 & \cellcolor{orange!50}0.77 & \cellcolor{yellow!50}0.53 & \cellcolor{yellow!50}0.4 & \cellcolor{yellow!50}1.07 & 0.81 & 0.87 & \cellcolor{red!50}1.21 & \cellcolor{yellow!50}1.14 & \cellcolor{red!50}0.47 & 0.70 & 1.36 & \cellcolor{yellow!50}0.36 & \cellcolor{yellow!50}0.58 & \cellcolor{orange!50}0.43 & \cellcolor{orange!50}0.75 & \cellcolor{red!50}0.07h \\
\hline
\end{tabular}}
\caption{Quantitative evaluation of 3D geometry precision on the DTU dataset.}
\vspace{-0.3cm}
\label{tab:dtu}
\end{table}

\noindent\textbf{3D Geometry precision:}
Table~\ref{tab:dtu} presents a quantitative comparison between our approach and other methods aimed at improving 3D reconstruction accuracy. Our approach not only surpasses other explicit methods in geometric precision but also competes favorably with SDF-based implicit methods. Note that compared with other implicit methods, our approach also benefits from faster training speeds, thereby increasing its practical applicability in real-world scenarios. 
Figure~\ref{fig:mesh} presents a qualitative comparison of mesh reconstruction using 3DGS, SuGaR, and our method on DTU scenes 24, 106, and 122. SuGaR tends to generate flat disks due to its regularization approach; however, these disks are not always perfectly aligned with the underlying geometry. NeuS employs a signed distance function, resulting in smoother surfaces, but could also lose high-frequency details due to strong smoothness priors. In contrast, our method maintains a balance between smoothness and detail preservation.

\begin{figure}[htbp]
    \centering
    \begin{tabular}{cccc}
     \includegraphics[width=0.2\textwidth]{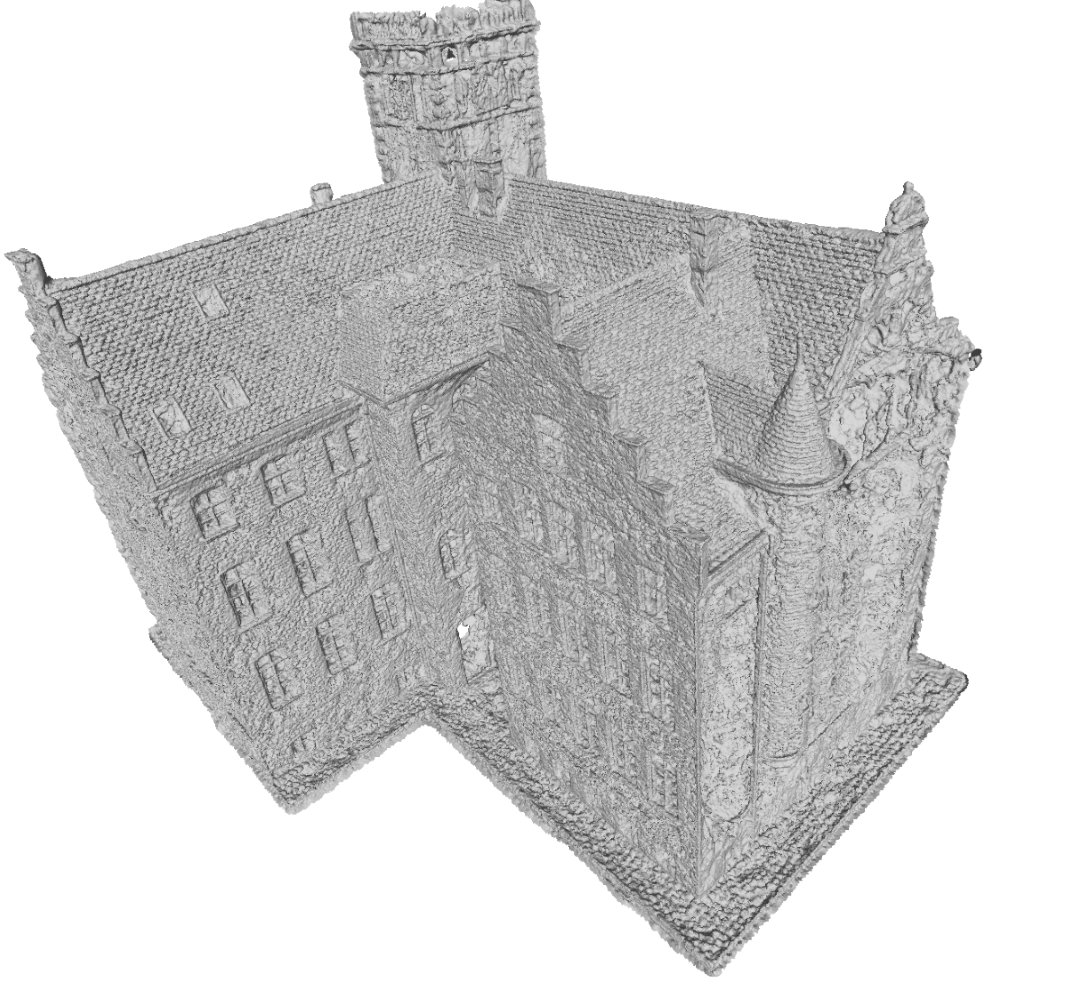}   
     &\includegraphics[width=0.2\textwidth]{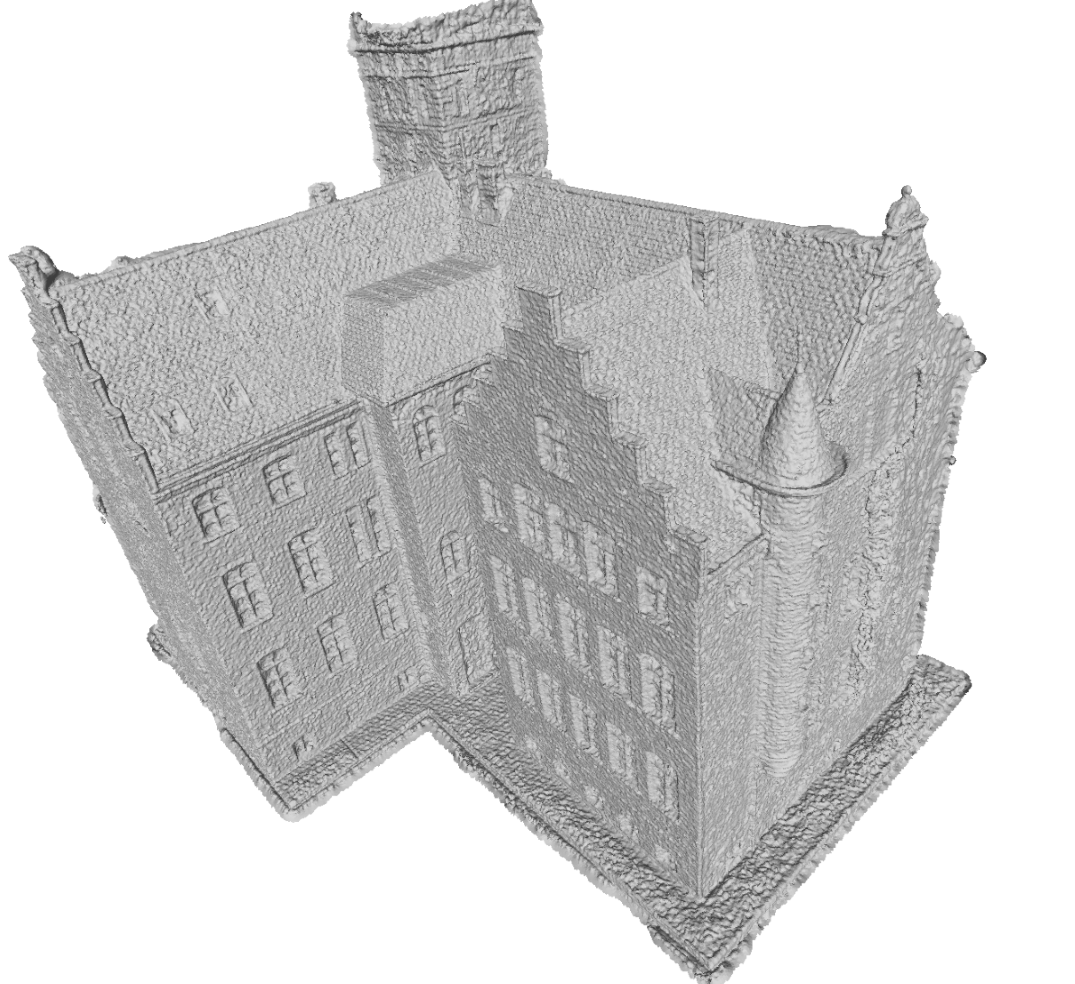}   
     &\includegraphics[width=0.2\textwidth]{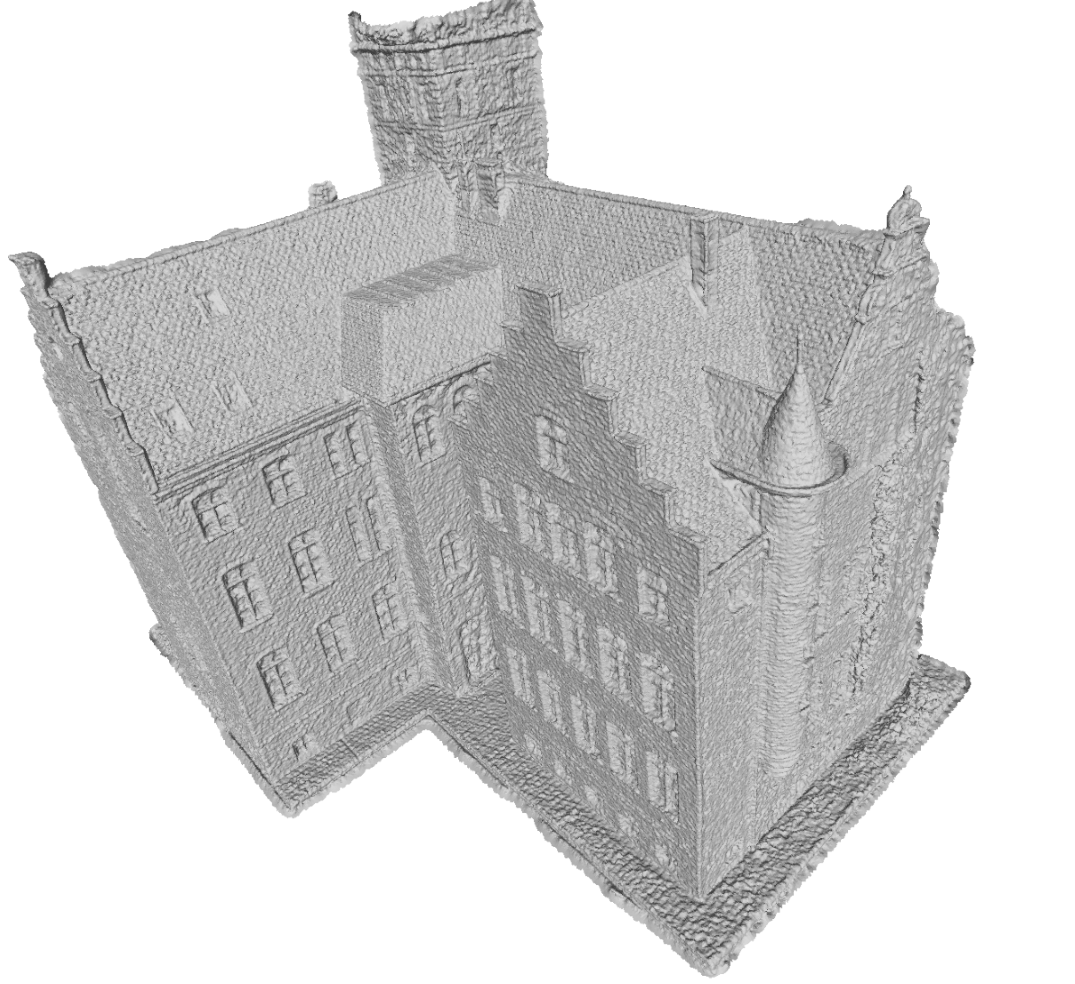} 
     &\includegraphics[width=0.2\textwidth]{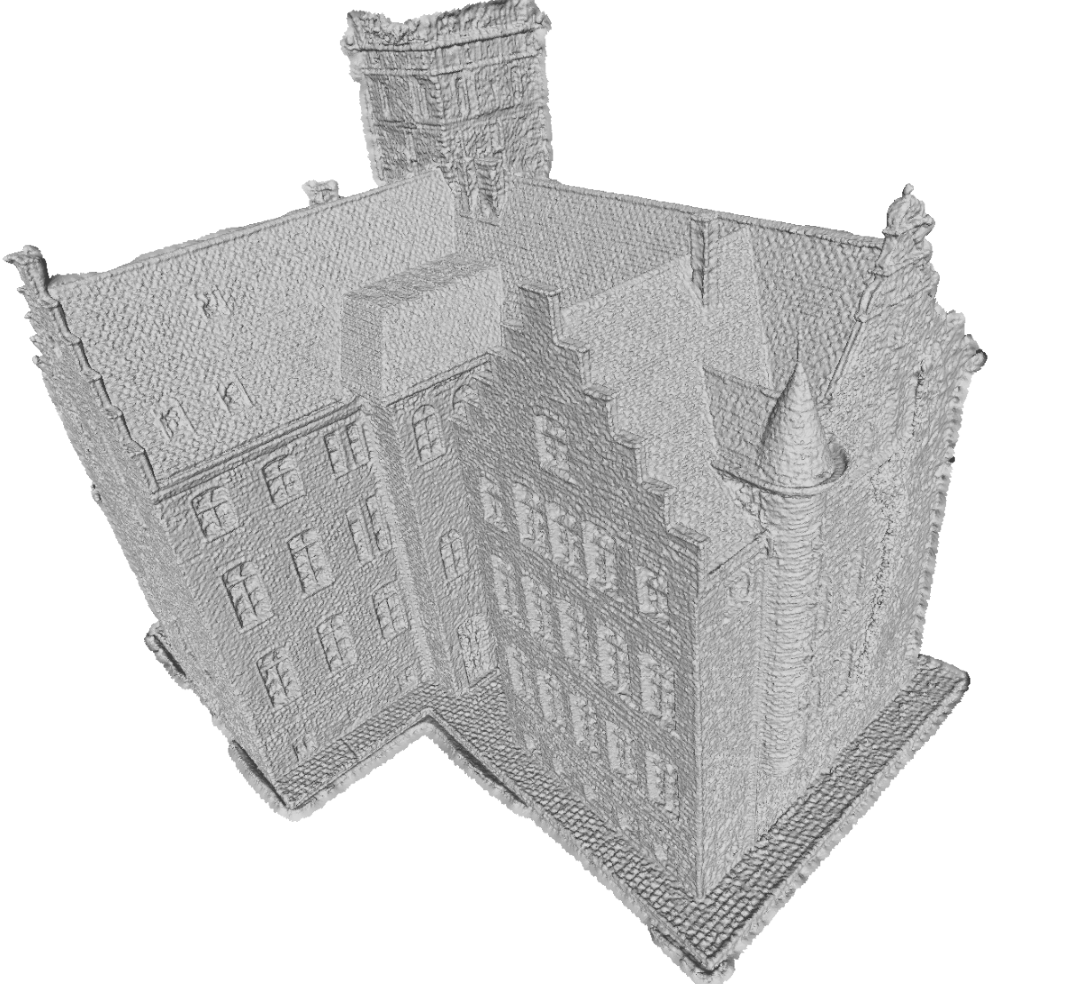} \\
     (a) No $atomization$ & (b) No $\mathcal{L}_{normal}$ & (c) No $\mathcal{L}_{ms-ssim}$ & Full model
    \end{tabular}
    \caption{Qualitative Results of Ablation on the DTU scene 24.}
    \vspace{-0.5cm}
    \label{fig:Qualitative Results of Ablation}
\end{figure}

\noindent\textbf{Ablation Study:}
We conducted ablation studies on the Tanks \& Temples and DTU datasets~\cite{Knapitsch2017, aanaes2016large} to evaluate the impact of different configurations on our model's performance. The first configuration involved the removal of Atomized Proliferation from our model. Following that, we kept Atomized Proliferation but discarded either of the loss functions during training. The results from these configurations and the full model are shown in Table~\ref{tab:ablation} and Figure~\ref{fig:Qualitative Results of Ablation}.
The results indicate that Atomized Proliferation is crucial to our model's performance. Excluding Atomized Proliferation resulted in the most significant performance decline. This suggests that simply incorporating geometry regularization in the loss functions tends to merely approximate the scene with elongated Gaussian ellipsoids instead of aligning accurately with the underlying geometry. Additionally, omitting either $\mathcal{L}_{normal}$ loss or  $\mathcal{L}_{ms-ssim}$ loss also leads to performance degradation, but to a lesser extent. The full model, which incorporates all components, consistently achieves the best performance, indicating that each element contributes to the overall effectiveness of the model.

\begin{table}[H]
\centering
    \resizebox{0.5\linewidth}{!}{
    \begin{tabular}{c|ccc|c}
        \hline
        & \multicolumn{3}{c|}{Tanks\&Temples} & DTU\\
        & $\text{PSNR}^\uparrow$ & $\text{SSIM}^\uparrow$ & $\text{LPIPS}^\downarrow$ & $\text{Chamfer-Distance}^\downarrow$\\
        \hline
        No $atomization$ & 23.25 & 0.815 & 0.228&1.23 \\
        No $\mathcal{L}_{normal}$ & 23.58 & 0.840 & 0.181&0.82 \\
        No $\mathcal{L}_{ms-ssim}$ & 23.58 & 0.837 & 0.185&0.78 \\
        Full model & \cellcolor{red!50}23.70 & \cellcolor{red!50}0.849 & \cellcolor{red!50}0.166&\cellcolor{red!50}0.75 \\ \hline
    \end{tabular}}
    \caption{Quantitative Results of Ablation on the Tanks\&Temples and DTU datasets.}
    \vspace{-1cm}
    \label{tab:ablation}
\end{table}

%% file: sec/5_conclusion.tex
\section{Conclusion}
\vspace{-0.2cm}
In this paper, we introduced AtomGS, an approach that enhances radiance field reconstruction by focusing on uniform densification through Atomized Proliferation and refining surface details via Geometry-Guided Optimization. Our approach significantly reduces noisy geometry and blurry artifacts that are common in the previous 3DGS methods. 

Nonetheless, AtomGS has its own limitations. Similar to the previous methods, our method may not produce accurate geometry for highly specular or semi-transparent material. While our method in general requires fewer GS primitives than the original 3DGS method to achieve improved visual quality, our proliferation strategy could sometimes produce more GS primitives to represent all details for highly complex environments. In the future, we aim to develop an improved pruning or merging strategy to achieve a more compact result.

%% file: sup.tex
\appendix





\section{Atomized Proliferation Algorithm}

The Atomized Proliferation algorithm is summarized in Algorithm~\ref{alg}. It starts with setting parameters for Clone Threshold ($\tau_c$), Split Threshold ($\tau_s$), Prune Threshold ($\epsilon$), Atom Scale ($\mathcal{S}$), and defining duration limits for atomized proliferation ($t_a$) and warm-up phase ($t_w$). The algorithm iteratively processes each Gaussian property ($\bm{\mu}, \bm{\Sigma}, \bm{c}, \alpha$) from the Gaussian set ($\bm{M}, \bm{S}, \bm{C}, \bm{A}$). A Gaussian is pruned if its $\alpha$ falls below the threshold $\epsilon$ or its covariance ($\bm{\Sigma}$) is excessively large. If the gradient of the loss ($\nabla_p \mathcal{L}$) exceeds $\tau_c$, the Gaussian is cloned to potentially bridge geometry gaps. Additionally, the Gaussian is split when $\nabla_p \mathcal{L}$ meets a dynamically adjusted threshold based on the warm-up progress and if the norm of $\bm{\Sigma}$ exceeds the Atom Scale ($\mathcal{S}$). Atomization takes place when the minimum norm of $\bm{S}$ is less than or equal to Atom Scale $\mathcal{S}$ and within the proliferation timeframe ($t_a$), ensuring detail refinement before the proliferation endpoint.

\begin{algorithm}
\caption{Atomized Proliferation}
\begin{algorithmic}[1]
    \REQUIRE Clone Threshold $\tau_c$, Split Threshold $\tau_s$, Prune Threshold $\epsilon$,
    \\ \qquad\; Atom Scale $\mathcal{S}$, Atomized Proliferation until $t_a$, Warm-Up until $t_w$

    \FOR{\textbf{all} Gaussian$(\bm{\mu},\bm{\Sigma},\bm{c},\alpha)$ in $(\bm{M},\bm{S},\bm{C},\bm{A})$}
        \IF {$\alpha < \epsilon$ or IsTooLarge$(\bm{\Sigma})$}
            \STATE $\text{Prune$(\bm{\mu},\bm{\Sigma},\bm{c},\alpha)$}$
        \ENDIF
        \IF {$\nabla_p \mathcal{L}\geq \tau_c$}
            \STATE $\text{Clone$(\bm{\mu},\bm{\Sigma},\bm{c},\alpha)$}$
        \ENDIF
        \IF {$\nabla_p \mathcal{L}\geq \min\left(\frac{i}{t_w}\tau_s, \tau_s\right)$ and $||\bm{S}||_{\max} > \mathcal{S}$}
            \STATE $\text{Split$(\bm{\mu},\bm{\Sigma},\bm{c},\alpha)$}$
        \ENDIF
        \IF {$||\bm{S}||_{\min} \leq \mathcal{S}$ and $i<t_a$}
            \STATE $\text{Atomize$(\bm{\mu},\bm{\Sigma},\bm{c},\alpha)$}$
        \ENDIF    
    \ENDFOR
\end{algorithmic}
\label{alg}
\end{algorithm}

\section{Gaussian Proliferation Trend}
\begin{figure}[!h]
\centering
  \includegraphics[width=\textwidth]{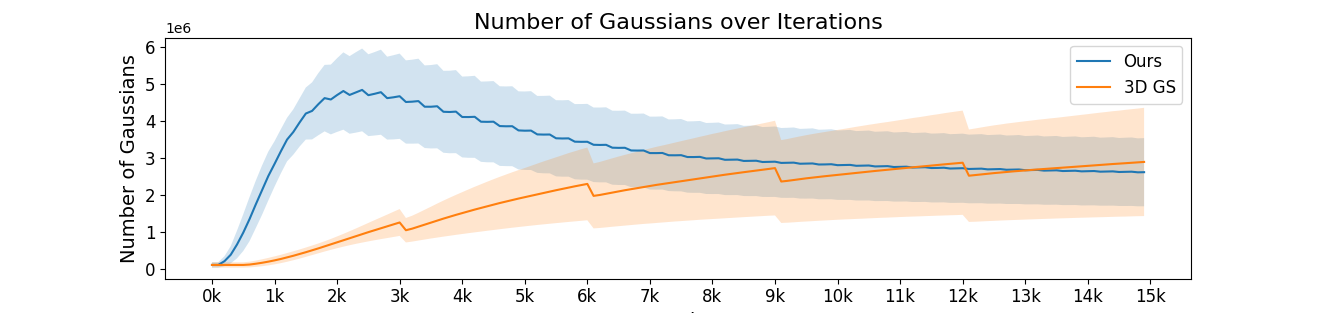} 
    \caption{Gaussian Proliferation Trend.}
    \label{fig:proliferation_trend}
\end{figure}

In Figure~\ref{fig:proliferation_trend}, we illustrate the Gaussian Proliferation Trend, which tracks the count of Gaussians across iterations for nine different scenes within the Mip-NeRF360 dataset. The depicted curve represents the average number of Gaussians across these scenes, with the curve's width indicates the standard deviation. The fluctuations observed highlight the effectiveness of the opacity resetting strategy in eliminating redundant Gaussians. Initially, the 3DGS method struggles to densify Gaussians, as indicated by an increasing standard deviation, and it appears unable to stabilize by the end of the proliferation stage. In contrast, our method employs a warm-up strategy that aggressively densifies Gaussians at the initial stage, followed by a phase where Gaussians begin to merge, leading to a declining and stabilizing trend in Gaussian proliferation.

On the Mip-NeRF360 dataset, our method demonstrates efficiency with an average training time of 0.28 hours and a final model size of 749MB, compared to 3DGS, which takes 0.40 hours for training and results in a model size of 869MB. This indicates that our approach achieves superior quality without compromising on training time or model size.

\section{Implementation Details}
\noindent\textbf{Codebase:} 
We have developed AtomGS based on the 3D Gaussian Splatting (3DGS) framework~\cite{kerbl20233d}.
To facilitate Edge-Aware Normal Loss computation and Poisson mesh extraction, we have implemented an additional feature renderer. This renderer generates various maps, including accumulation, median and mean depth, normal, and curvature maps. Additionally, we've developed an interactive real-time viewer that allows for the monitoring of these features, providing a detailed analysis of Gaussians in terms of both RGB and geometric information. For a detailed derivation of these implementations, please refer to Section~\ref{rendering}.

\noindent\textbf{Hyper Parameter Settings:}
Following the 3DGS, we set the Clone Threshold at $\tau_c=0.002$, Split Threshold at $\tau_s=0.002$, and Prune Threshold at $\epsilon=0.005$. For the Atom-related settings, we set Atom Scale at the first percentile of distances from the input SfM points $\mathcal{S}=P_1(\bm{d})$, Atomized Proliferation until iteration at $t_a=7000$, and Warm-Up until iteration at $t_w=7000$. For optimization, the weights for MS-SSIM and normal loss calculations are both set at $\lambda_{ms-ssim}=\lambda_{normal}=0.1$. When working with object-centered datasets that lack extensive backgrounds, we advise setting the scale learning rate $\eta_s=0$ to maximize geometric accuracy. Specifically for the DTU dataset, we set Prune Threshold $\epsilon=0.5$, loss weight $\lambda_{ms-ssim}=1$, Atom Scale $\mathcal{S}=P_{10}(\bm{d})$, ant total training iterations at 7k.

\begin{figure}[!h]
\centering
  \includegraphics[width=0.9\textwidth]{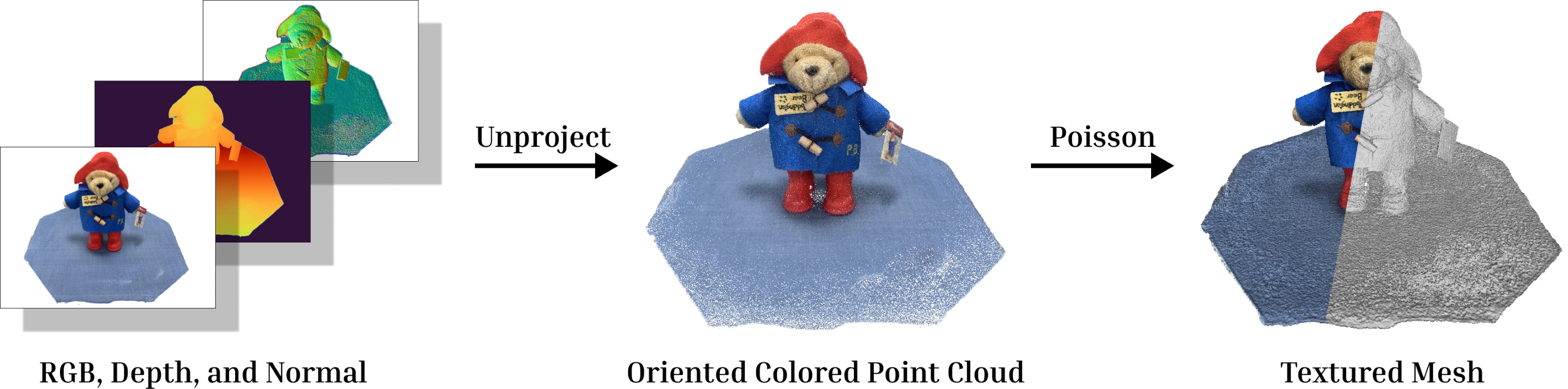} 
    \caption{Poisson Mesh Extraction.}
    \label{fig:mesh_extraction}
\end{figure}

\noindent\textbf{Mesh Extraction:}
Mesh extraction involves rendering depth maps from training views, which use median depth values from splats projected onto pixels. These maps are then converted back into 3D space to derive corresponding normal maps. The oriented colored point cloud generated from the RGB image, depth map, and normal map serves as the input for the Poisson extraction method~\cite{kazhdan2006poisson}, which is used to create the textured mesh. This process is illustrated in Figure~\ref{fig:mesh_extraction}.

\noindent\textbf{Hardware:} 
All experiments are conducted on a single NVIDIA GeForce RTX 4090 GPU.

\section{Gaussian Splatting and Additional Feature Rendering}
\label{rendering}

\noindent\textbf{Splatting:}
During this stage, 3D Gaussians are projected into the 2D image space to facilitate rendering. Utilizing the viewing transformation $\bm{W}$ and the 3D covariance matrix $\bm{\Sigma}$, the projected 2D covariance matrix $\bm{\Sigma}^\prime$ is computed through $\bm{\Sigma}^\prime = \bm{JW\Sigma W^\top J^\top}$.
Additionally, we can use the same transformation to compute $\bm{\mu}^\prime \in \mathbb{R}^2$ in 2D projected space. Given the position of a pixel $\bm{x}\in \mathbb{R}^2$, 3D Gaussian splitting can be formed as follows:
\begin{equation}
    \mathcal{G}_i(\bm{x}) := \exp{(-\frac12(\bm{x}-\bm{\mu}_i^\prime)^\top\bm{\Sigma}_i^{\prime-1}(\bm{x}-\bm{\mu}_i^\prime))}.
\end{equation}

\noindent\textbf{Rendering:}
Upon receiving the position of a pixel $\bm{x}$, the distances to all overlapping Gaussians are computed using the viewing transformation $\bm{W}$, thereby generating a sorted list of Gaussians $\mathcal{N}:=\{G_1,...,G_N\}$. Subsequently, alpha compositing is employed to render accumulated weight for each pixel:
\begin{equation}
    w_i(\bm{x}) = \alpha_i\mathcal{G}_i(\bm{x})\prod_{j=1}^{i-1}(1-\alpha_j\mathcal{G}_j(\bm{x})).
\end{equation}
Using the above weight function, several key maps can be derived for each pixel:
\begin{enumerate}
\item RGB Color Map:
    \begin{equation}
    \bm{C}(\bm{x})=\sum_{i=1}^N c_iw_i(\bm{x})
\end{equation}
accumulates the RGB colors $c_i$, each weighted by the respective $w_i(\bm{x})$, to produce the final color output for each pixel.
    \item Accumulation Map:
    \begin{equation}
    \bm{A}(\bm{x})=\sum_{i=1}^N w_i(\bm{x}),
    \end{equation}
    which aggregates the computed weights across all Gaussians.
    \item Mean (Expected) Depth Map:
    \begin{equation}
    \bm{D}_{mean}(\bm{x})=\sum_{i=1}^N z_iw_i(\bm{x}),
\end{equation}
where $z_i$ represents the depth associated with each Gaussian, weighted by $w_i(\bm{x})$.
    \item Median Depth Map:
    \begin{equation}
    \bm{D}_{median}(\bm{x}) = z_i \quad \text{where} \quad i = \min_i \left\{  \prod_{j=1}^{i-1} (1-\alpha_j \mathcal{G}_j(\bm{x})) > 0.5 \right\}
\end{equation}
calculates the median depth by identifying the first Gaussian for which the Transmittance value exceeds 0.5.
\end{enumerate}

\noindent\textbf{Depth Map Unprojection:}
Given a depth map $\bm{D}^\prime$ of size $H \times W$, where $(i, j)$ are pixel coordinates and $d_{i,j}$ is the depth at pixel $(i, j)$, the unprojecting steps are as follows:
\begin{enumerate}
    
    \item \textbf{Normalization:} The pixel coordinates are normalized to the range $[-1, 1]$ using $x_{\text{norm}} = \frac{2j}{W-1} - 1$ and $y_{\text{norm}} = \frac{2i}{H-1} - 1$.
    
    \item \textbf{Homogeneous Coordinates in Camera Space:} The normalized coordinates are then transformed into homogeneous camera space coordinates $\mathbf{p}_{\text{camera}} = [x_{\text{norm}}, y_{\text{norm}}, d_{i,j}]^T$.
    
    \item \textbf{Depth Scaling:} Using elements $f_1$ and $f_2$ from the camera projection matrix $\mathbf{K}$, the depth values are scaled as $s_{d_{i,j}} = \frac{f_1 \cdot d_{i,j} + f_2}{d_{i,j}}$. The adjusted camera space coordinates are set to $\mathbf{p}_{\text{camera}}' = [x_{\text{norm}}, y_{\text{norm}}, s_{d_{i,j}}]^T$.
    
    \item \textbf{World Space Transformation:} The transformed camera space coordinates are then multiplied by the inverse of the full projection transform matrix $\mathbf{T}$, resulting in $\mathbf{p}_{\text{world}} = \mathbf{T} \times [x_{\text{norm}}, y_{\text{norm}}, s_{d_{i,j}}, 1]^T$.
    
    \item \textbf{Discarding Homogeneous Coordinate:} Finally, to obtain Cartesian coordinates, the homogeneous coordinate is discarded: $\mathbf{p}_{\text{world}}' = \frac{\mathbf{p}_{\text{world}}[0:3]}{\mathbf{p}_{\text{world}}[3]}$. This results in $\bm{D} = \mathbf{p}^\prime_{\text{world}}$ being the 3D coordinates in world space for each pixel.
\end{enumerate}

\noindent\textbf{Normal Map Calculation:}
Given an unprojected depth map, $\bm{D} \in \mathbb{R}^{H \times W \times 3}$, we can output the corresponding normal map using the cross product of the depth map's gradients:
\begin{equation}
    \bm{N} = \frac{\nabla_x\bm{D}\times\nabla_y\bm{D}}{||\nabla_x\bm{D}\times\nabla_y\bm{D}||}
\end{equation}

\section{Additional Results}
\begin{figure}[htbp]
    \centering
    \includegraphics[width=\textwidth]{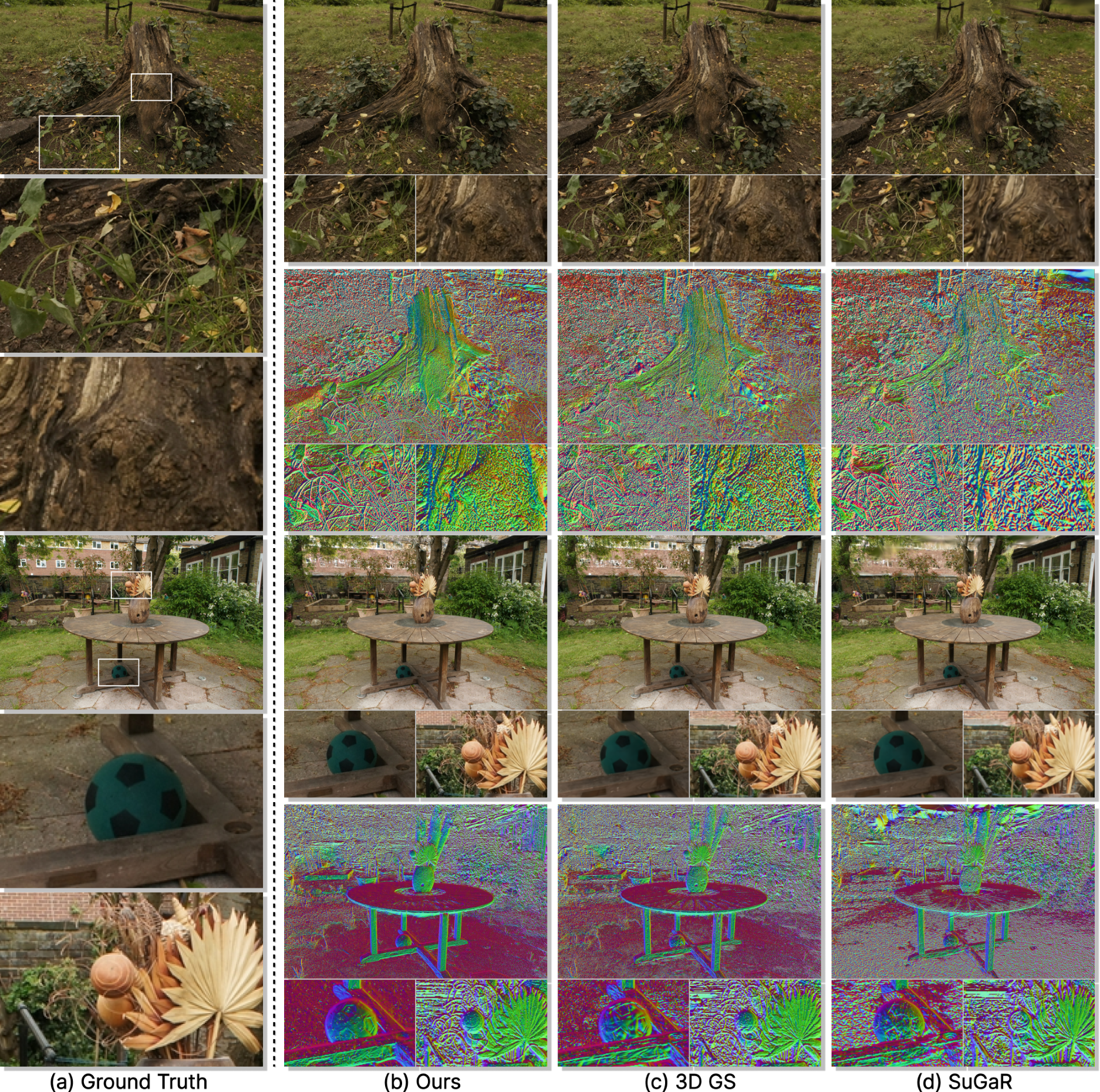}
    \caption{Radiance Field Comparison on the Mip-NeRF360 Dataset.}
    \label{fig:qualitative2}
\end{figure}

\begin{figure}[htbp]
    \centering
    \begin{tabular}{cccc}
         \rotatebox{90}{DTU 106}
         & \raisebox{-0.3\height}{\includegraphics[width=0.25\textwidth]{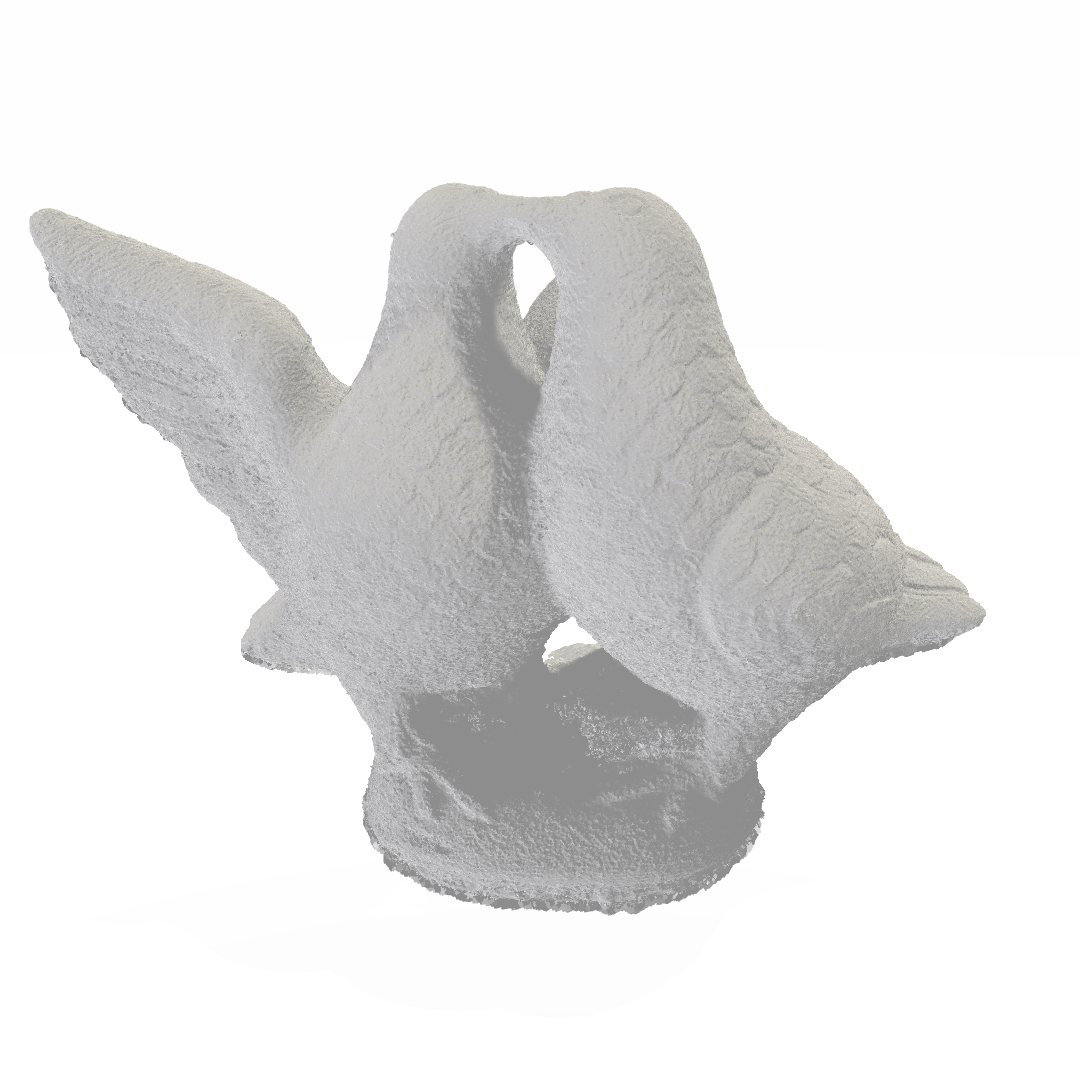}}
         & \raisebox{-0.3\height}{\includegraphics[width=0.25\textwidth]{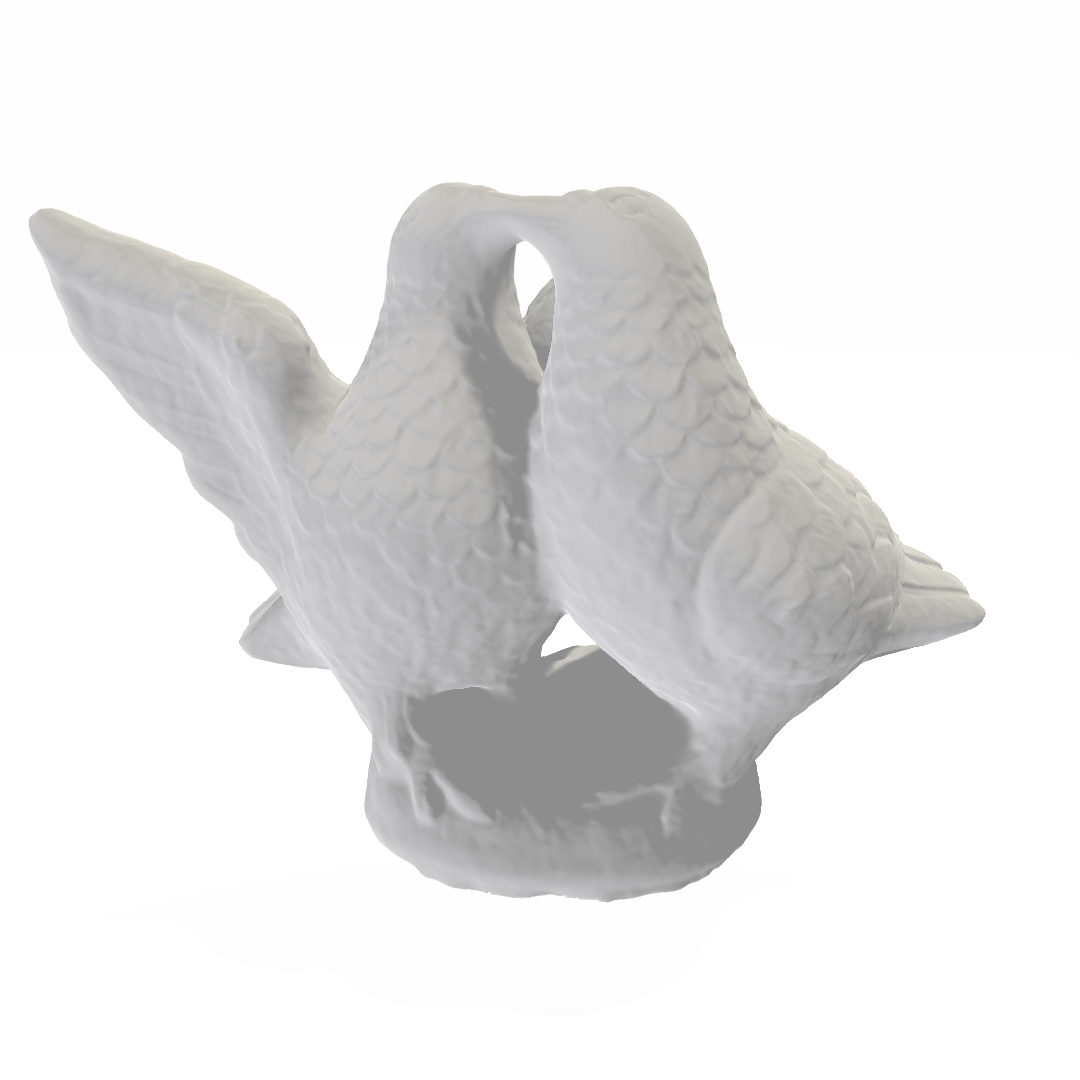}}
         & \raisebox{-0.3\height}{\includegraphics[width=0.25\textwidth]{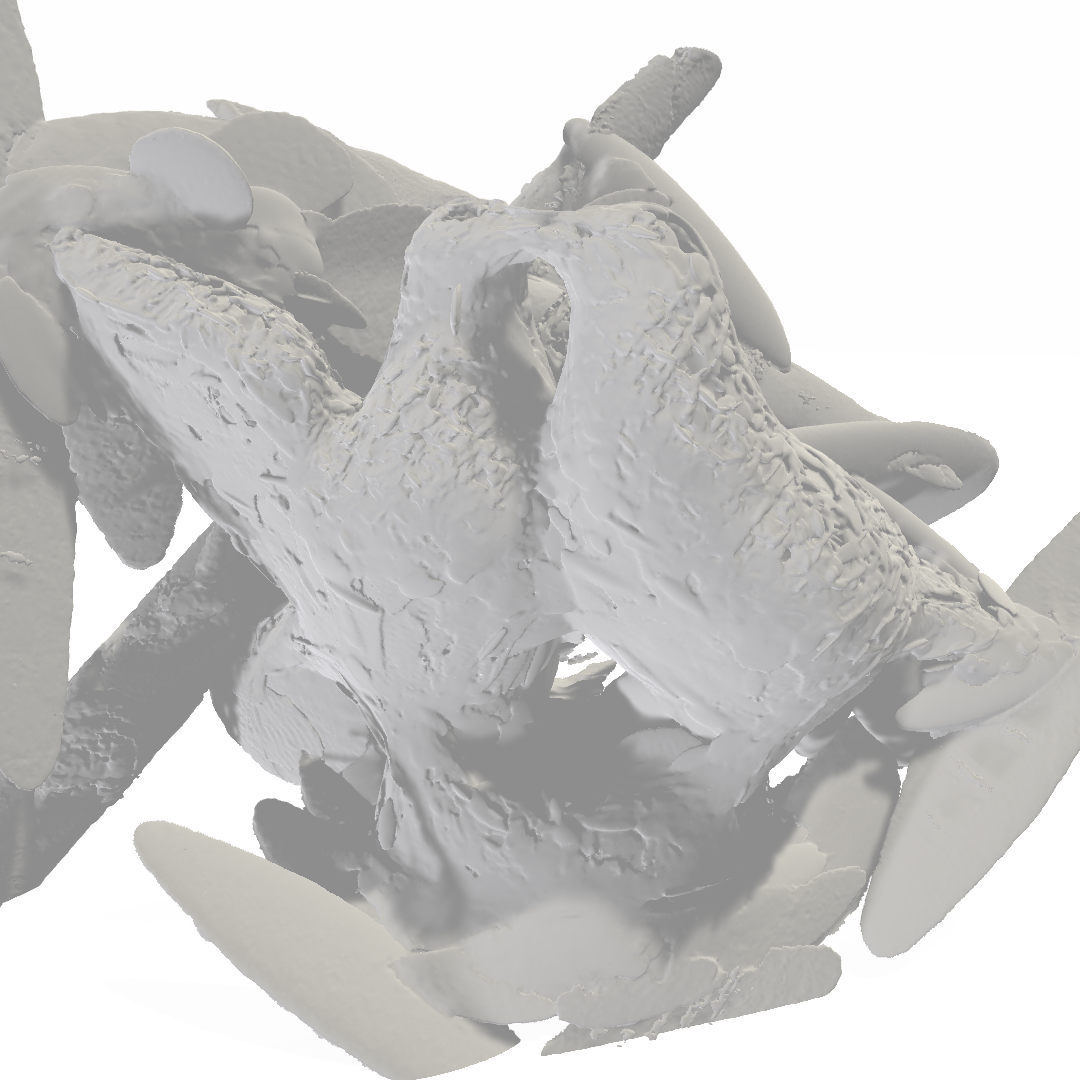}} \\
         \rotatebox{90}{DTU 122}
         & \raisebox{-0.3\height}{\includegraphics[width=0.25\textwidth]{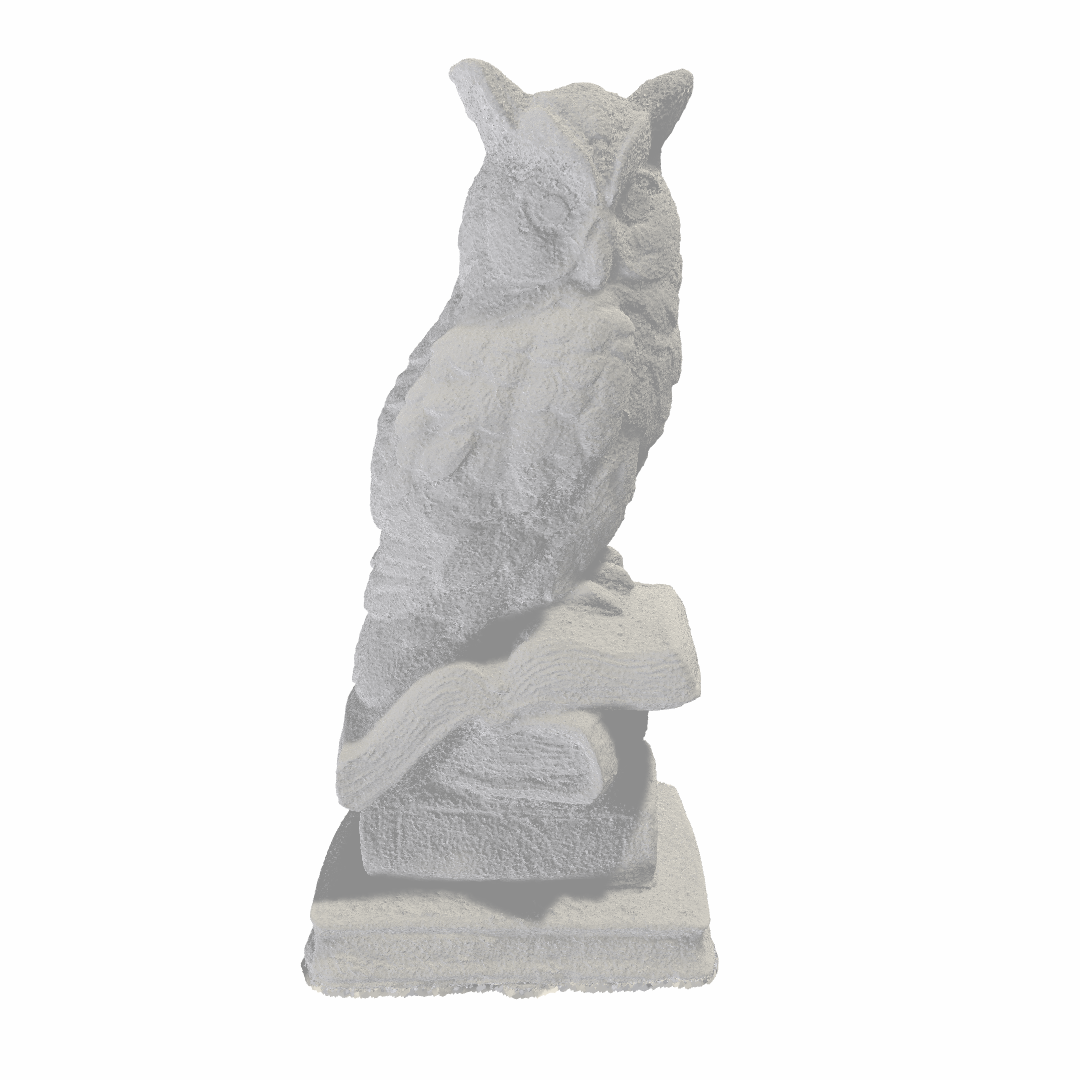}}
         & \raisebox{-0.3\height}{\includegraphics[width=0.25\textwidth]{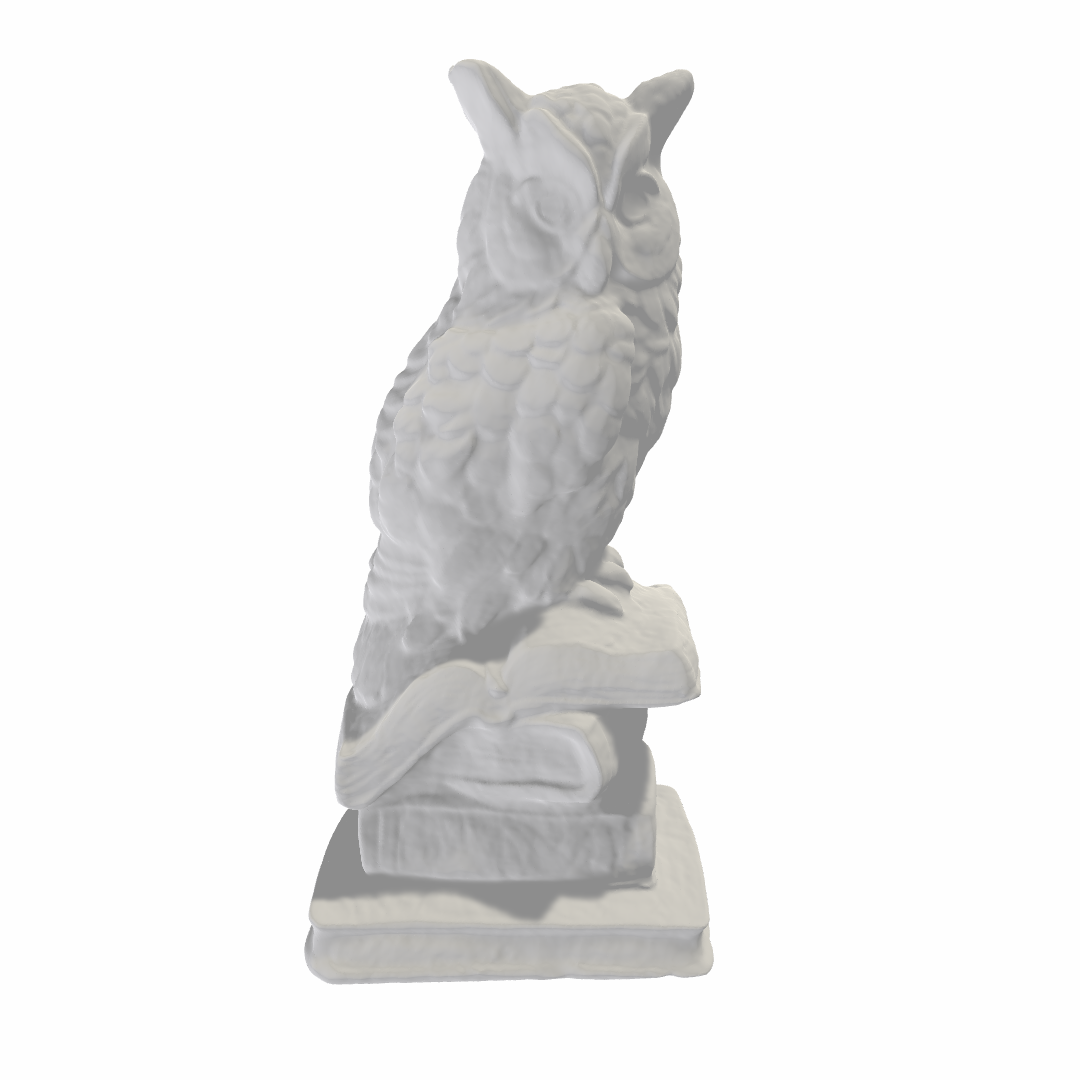}}
         & \raisebox{-0.3\height}{\includegraphics[width=0.25\textwidth]{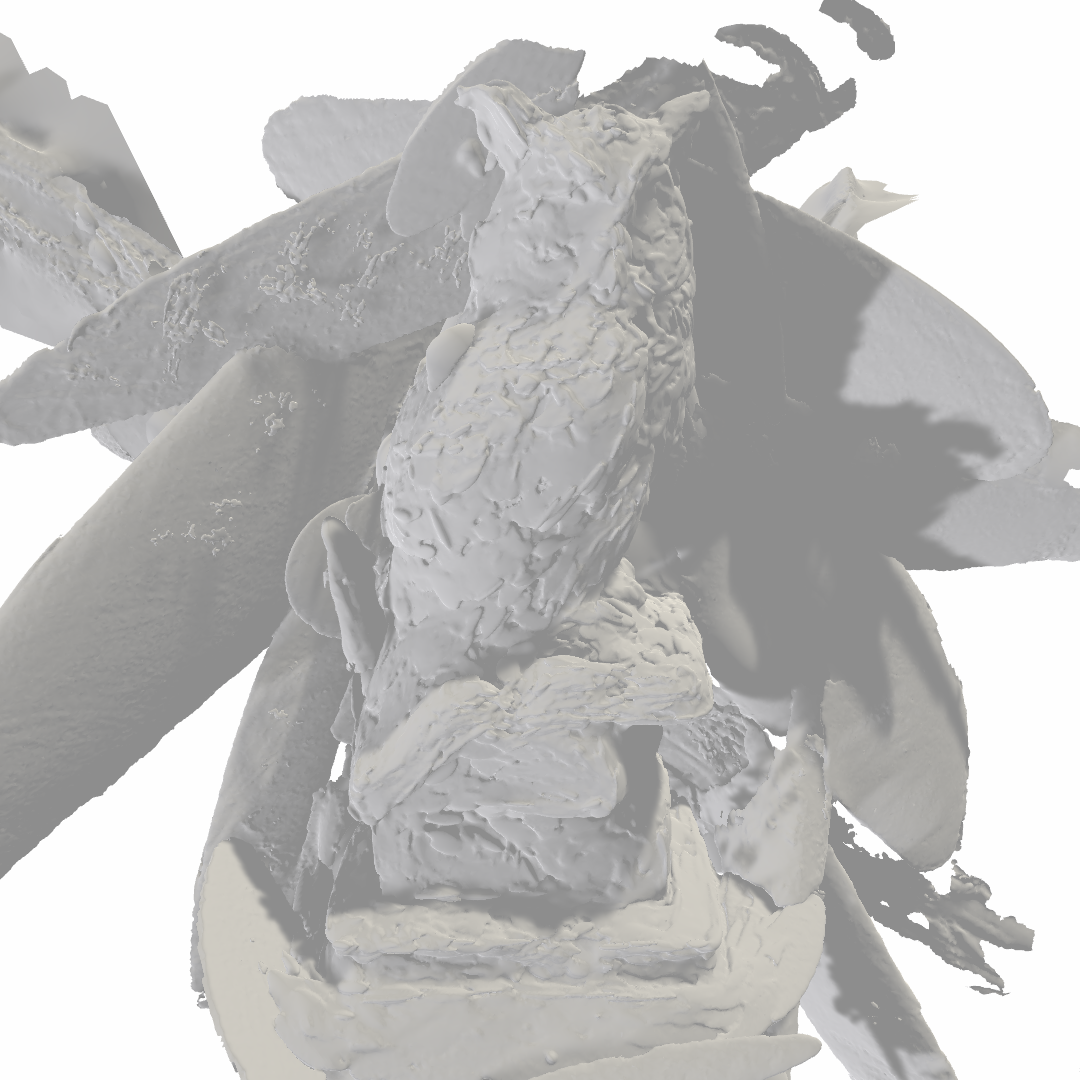}} \\
                  \rotatebox{90}{Lego}
         & \raisebox{-0.4\height}{\includegraphics[width=0.25\textwidth]{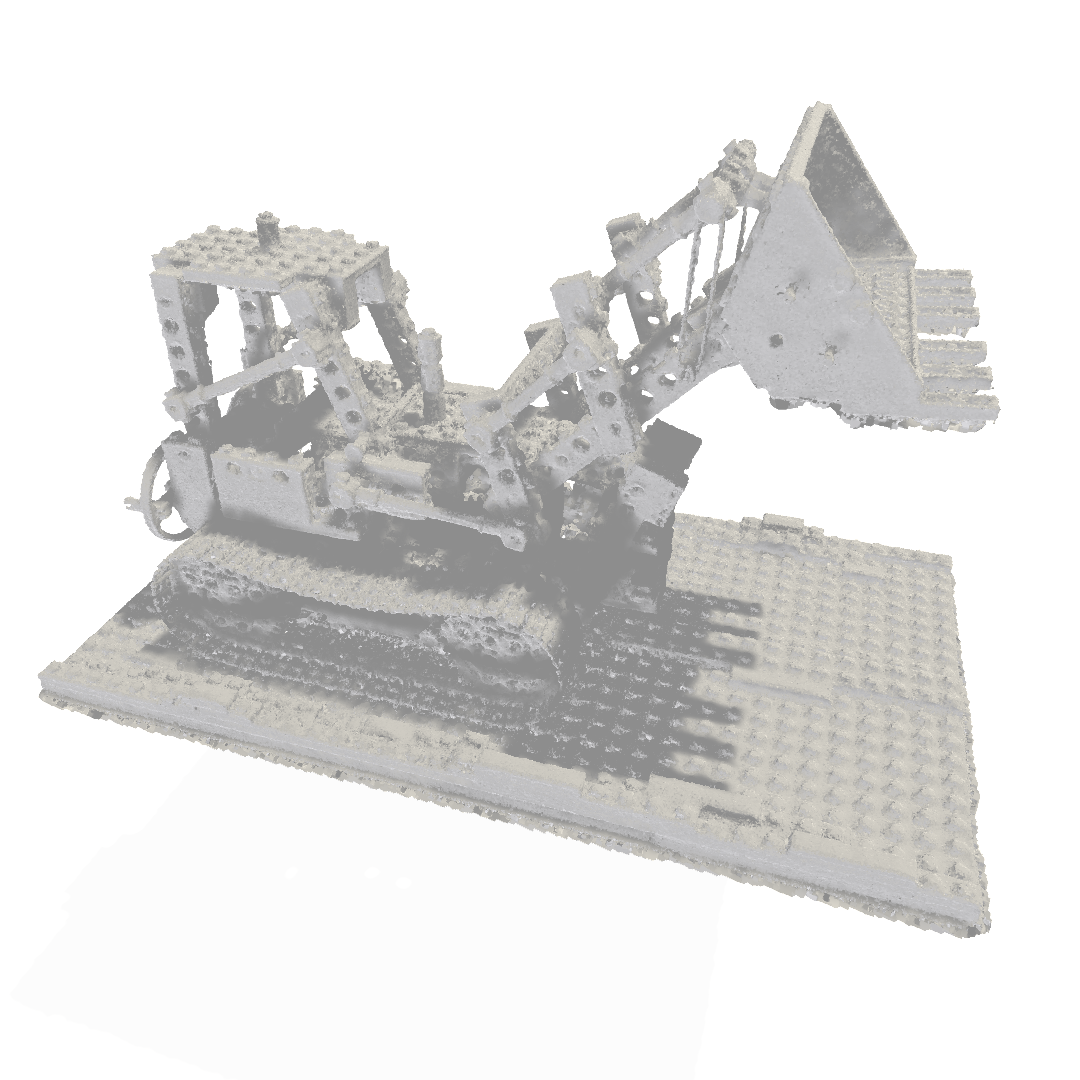}}
         & \raisebox{-0.4\height}{\includegraphics[width=0.25\textwidth]{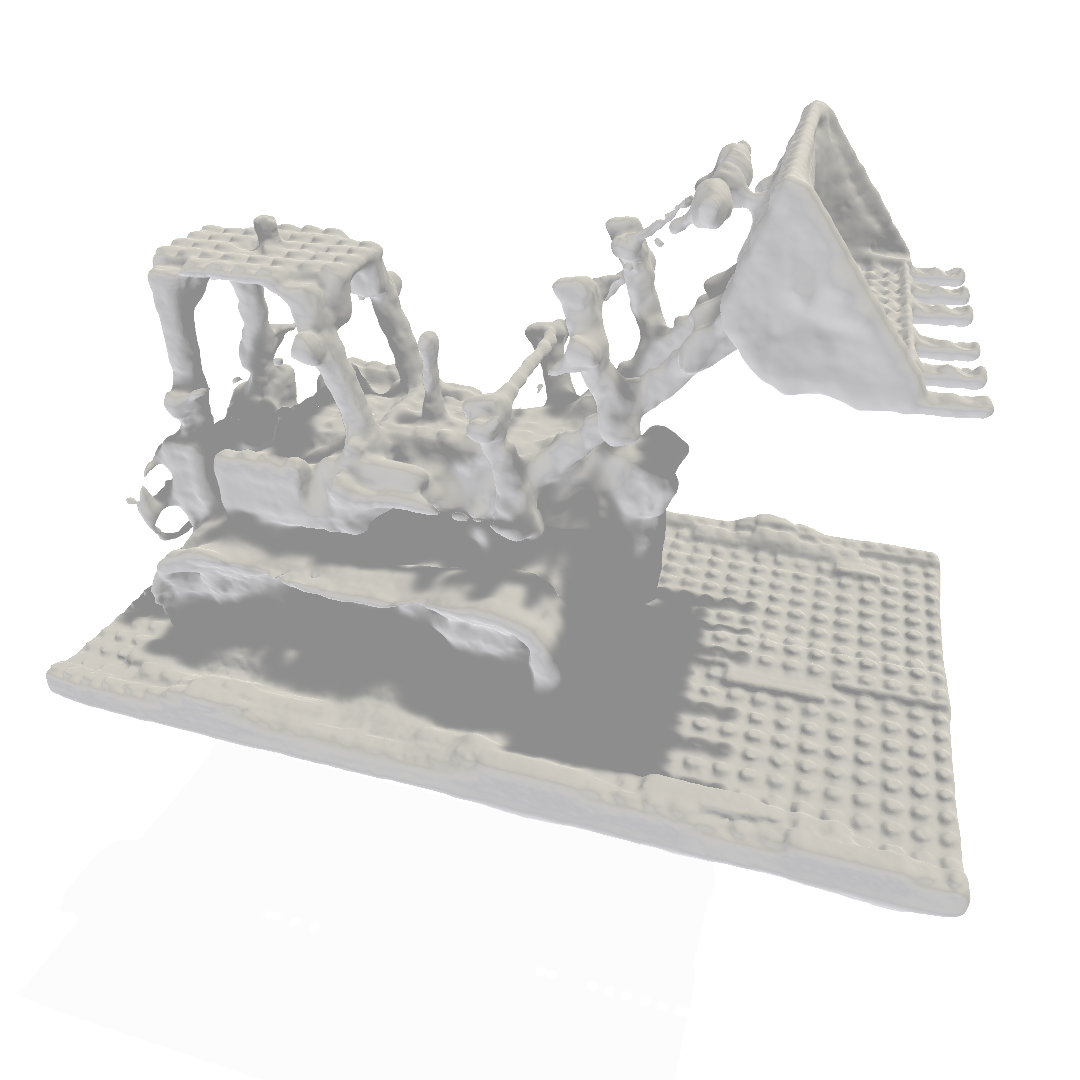}}
         & \raisebox{-0.4\height}{\includegraphics[width=0.25\textwidth]{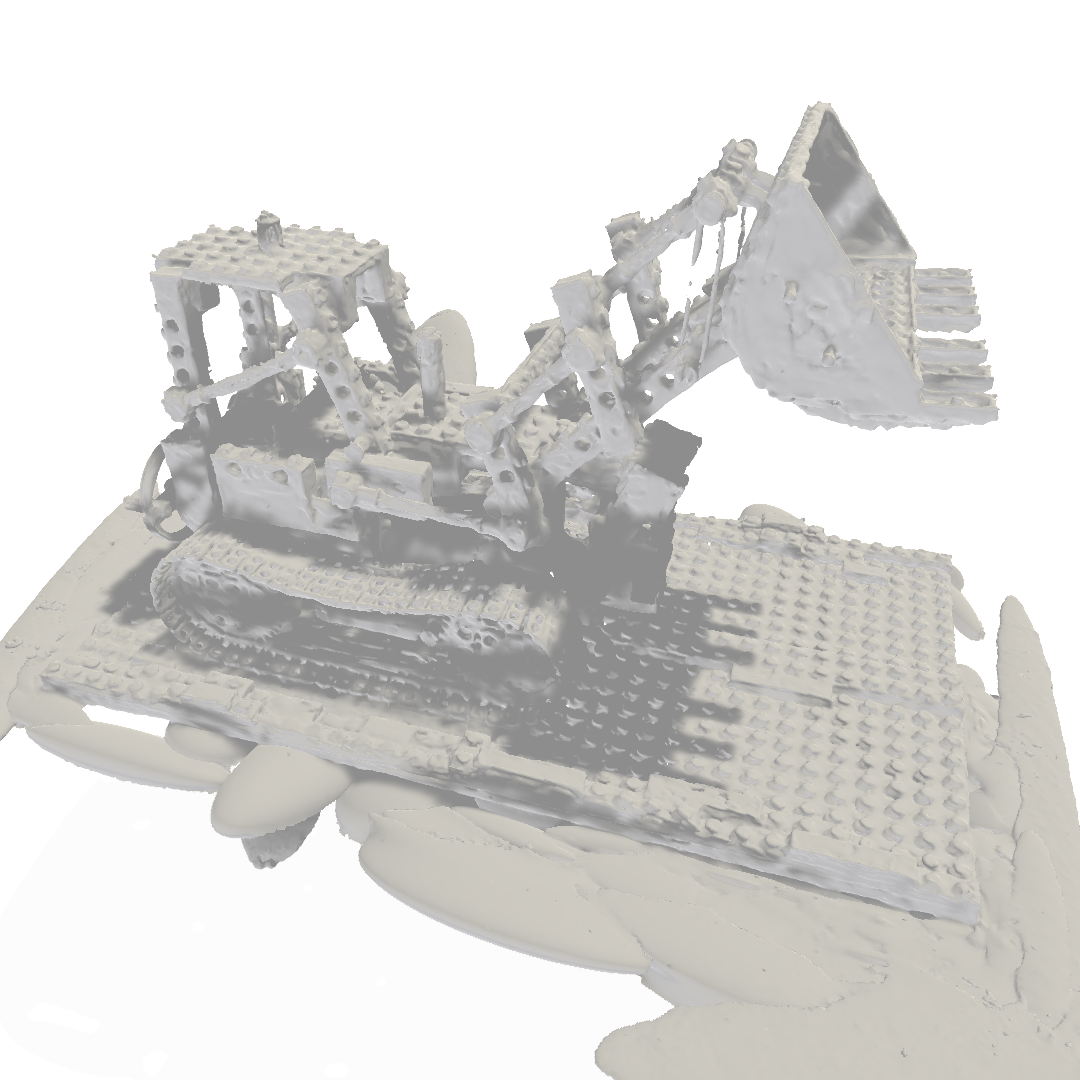}} \\
         \rotatebox{90}{Chair}
         & \raisebox{-0.4\height}{\includegraphics[width=0.25\textwidth]{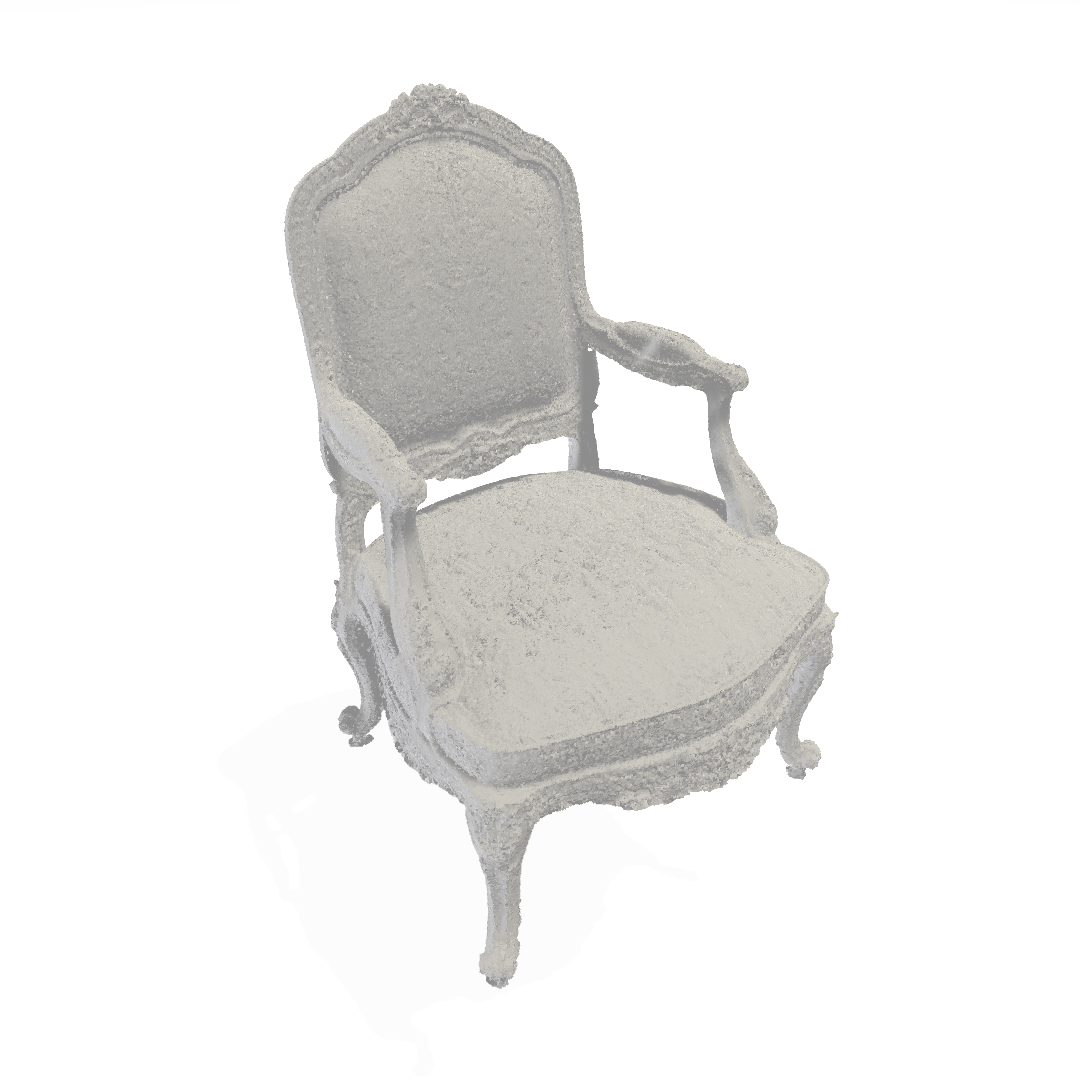}}
         & \raisebox{-0.4\height}{\includegraphics[width=0.25\textwidth]{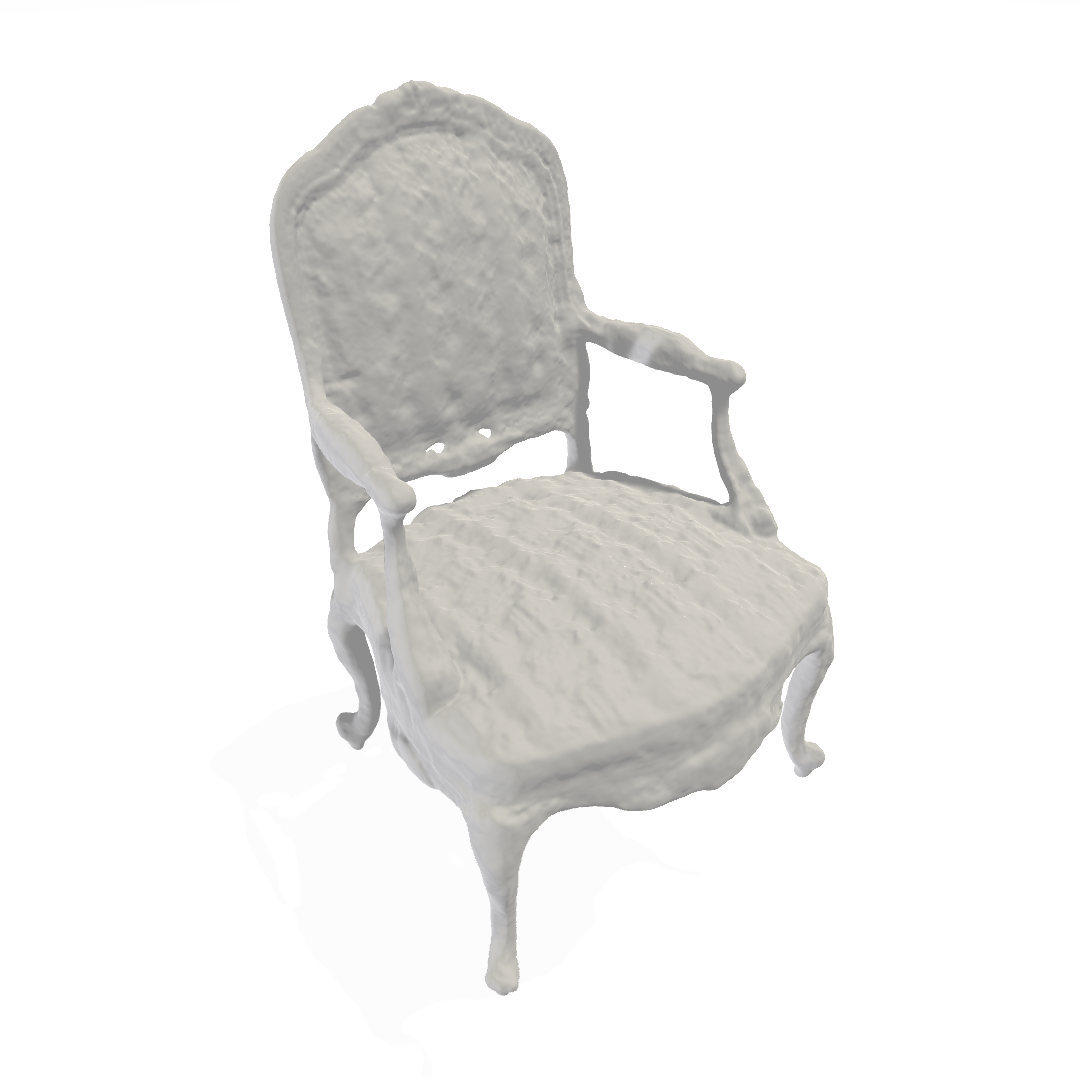}}
         & \raisebox{-0.4\height}{\includegraphics[width=0.25\textwidth]{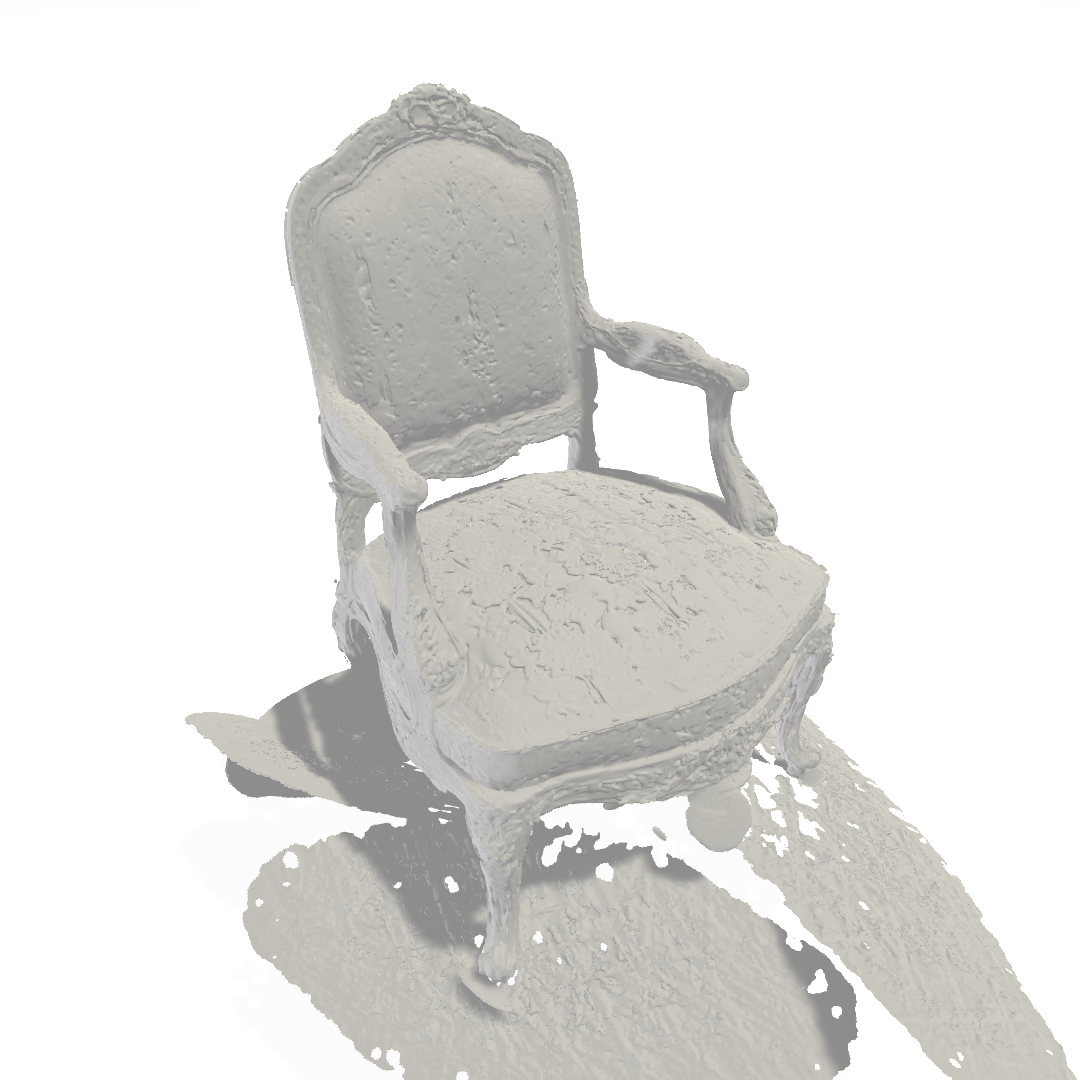}} \\
         & (a) Ours & (b) NeuS & (c) SuGaR \\
    \end{tabular}
    \caption{Mesh Comparison on the DTU and NeRF Synthetic Datasets~\cite{aanaes2016large, mildenhall2021nerf}}
    \label{fig:Mesh Comparison}
\end{figure}

We provide detailed per-scene metrics for the Mip-NeRF360 and Tank \& Temple datasets in Table~\ref{breakdown}. Additionally, we offer further insights through 2D rendering comparisons in Figure~\ref{fig:qualitative2} and 3D mesh comparisons in Figure~\ref{fig:Mesh Comparison}.

\begin{table}[htbp]
\centering
\resizebox{0.9\linewidth}{!}{
\begin{tabular}{cccccccccccccc}
\hline
     && Bicycle & Flowers & Garden & Stump & Treehill & Room & Counter & Kitchen & Bonsai & Truck & Train & Mean\\\hline
\multirow{3}{*}{\rotatebox[origin=c]{90}{PSNR}}
&3DGS & 25.10 & 21.52 & 27.18 & 26.49 & 22.37 & 31.22 & 28.96 & 30.98 & 32.18 & 25.39 & 22.02 &26.67\\
&SuGaR       & 23.13   & 19.67   & 25.30  & 24.23  & 21.44    & 29.85  & 27.53   & 29.33   & 30.47  & 22.69  & 20.47 &24.92 \\ 
&Ours            & 25.33            & 21.71            & 27.44           & 26.58          & 22.11             & 31.30         & 28.97            & 30.88            & 32.15           & 25.47          & 21.93         &26.72 \\
\hdashline
\multirow{3}{*}{\rotatebox[origin=c]{90}{SSIM}}
&3DGS & 0.763 & 0.603 & 0.860 & 0.763 & 0.626 & 0.916 & 0.905 & 0.923 & 0.939 & 0.878 & 0.812 &0.817\\
&SuGaR       & 0.663  & 0.514  & 0.793 & 0.669 & 0.558   & 0.901 & 0.882  & 0.892  & 0.928 & 0.827 & 0.762 &0.763\\
&Ours            & 0.772            & 0.611            & 0.865           & 0.774          & 0.633             & 0.918         & 0.906            & 0.925            & 0.938           & 0.880          & 0.817&0.822
\\
\hdashline
\multirow{3}{*}{\rotatebox[origin=c]{90}{LPIPS}}
&3DGS & 0.205 & 0.332 & 0.107 & 0.213 & 0.326 & 0.219 & 0.200 & 0.127 & 0.204 & 0.148 & 0.208 &0.208\\
&SuGaR      & 0.307  & 0.378  & 0.182 & 0.307 & 0.408   & 0.2395 & 0.222  & 0.167  & 0.211 & 0.175 & 0.259 &0.260\\ 
&Ours           & 0.203            & 0.325            & 0.104           & 0.202          & 0.317             & 0.222         & 0.202            & 0.127            & 0.202           & 0.133          & 0.200&0.203
          \\ \hline

\hline
\end{tabular}}
\caption{$\text{PSNR}^\uparrow$, $\text{SSIM}^\uparrow$, $\text{LPIPS}^\downarrow$ metrics for Mip-NeRF360 and Tank\&Temple datasets}
\label{breakdown}
\end{table}


In Figure~\ref{fig:qualitative2}, the RGB rendering results of our AtomGS show enhanced detail compared to those of Sugar. This is evident in the regions observed on the tree trunk in the "stump" scene and the slender, curly hay near the dried grass ornament in the "garden" scene, as highlighted in the magnified areas. While AtomGS's RGB renderings appear visually similar to those of 3DGS, the normal maps reveal that AtomGS better preserves geometry accuracy, such as the tree trunk in the "stump" scene and both the ground beneath the table and the surface of the soccer ball in the "garden" scene.

In Figure~\ref{fig:Mesh Comparison}, Neus, which uses Signed Distance Functions (SDF), produces the smoothest surfaces.  However it sometimes sacrifices sharp features, leading to overly smoothed surfaces. SuGaR attempts to convert every Gaussian into 2D ellipsoidal disks, resulting in relatively smooth surfaces. However, the disks do not always align perfectly with the surfaces, creating noticeable disk-shaped artifacts and sometimes overfitting the background. In contrast, AtomGS achieves smooth surfaces while retaining detailed geometries.

